# What General Intelligence Requires: Non-Reducible Constraints Across Levels of Description

**Subhomoy Bakshi**

Lakshmikumaran and Sridharan Attorneys, New Delhi, India

Correspondence: subhomoy.bakshi@lkslaw.com

ORCID: https://orcid.org/0009-0009-2234-7105

## Abstract

General intelligence, of the kind that underwrites the full range of human cognitive achievement, is not a property of computational architecture alone. This paper advances a single thesis: the structural constraints on general intelligence occupy distinct levels of description and are mutually non-reducible, in the sense that the special-sciences tradition gives to that term. It follows that no single architectural advance, and no continuation of the scaling programme by itself, can produce artificial general intelligence (AGI), and that research programmes must be evaluated against the full constraint profile rather than against performance on any one benchmark. The thesis is developed through a method that reads general intelligence through four evidential lenses, AI systems research, anthropology, law, and economics, each anchored to a distinct level of description, supplemented by speculative fiction used as a disciplined heuristic in the context of discovery rather than the context of justification. Applying the method yields a taxonomy of twenty-three structural constraints organised into eight clusters; six are examined in depth and ordered as an ascending ladder of levels, with explicit bridges showing why progress at one level cannot carry to the next. The argument issues in five falsifiable predictions, each stated with a named benchmark family and a disconfirmation condition, converting a descriptive framework into a research programme with a longer horizon than the scaling hypothesis implies.

## 1. Introduction

The question this paper asks has a long lineage. Turing (1950) set the terms of the modern debate by setting aside the question whether machines can think, which he judged too ill-defined to settle, in favour of an operational test of indistinguishable performance; the reframing offered here runs in the opposite direction, turning from whether a machine can be said to think to what general intelligence, in any substrate, would structurally require. The question of whether artificial general intelligence is achievable, and if so under what conditions, has been addressed almost exclusively within the disciplines most proximate to its development: computer science, cognitive science, and AI safety research. This disciplinary concentration is itself a structural problem. General intelligence, the kind that evolved in human beings and that underpins the full range of human cognitive achievement, is not a property of computational architecture alone. It is simultaneously a product of evolutionary history, an operating condition defined by institutional and legal systems, and an economic participant embedded in incentive structures. A framework for evaluating what general intelligence requires must draw on all of these traditions at once, and it must do so in a way that keeps their distinct explanatory vocabularies visible rather than collapsing them into a single computational account.

The thesis of this paper is that the structural constraints on general intelligence live at distinct levels of description and are mutually non-reducible in the sense established by Fodor's (1974) analysis of the special sciences: the kinds picked out at one level are multiply realisable at the level below, so no bridge law reducing one to the other can be stated as a natural kind predicate. From this thesis two consequences follow directly. First, no single architectural advance can produce AGI, because an advance at one level does not, by itself, constitute progress at the others. Second, research programmes that claim to advance toward general intelligence must be evaluated against the full constraint profile, not against performance on the single dimension they target. The framework and the taxonomy developed below are instruments of this thesis. They exist to make the non-reducibility claim precise, to show where its evidence lies, and to render it testable.

The dominant research programme in AI over the past decade has operated under what is sometimes called the scaling hypothesis: the proposition that increases in model size, training data, and computational resources will, perhaps in combination with architectural refinements, yield systems approaching general intelligence (Brown et al., 2020; Hoffmann et al., 2022; Bubeck et al., 2023; Morris et al., 2024). The empirical record of this programme has been striking; systems trained on large corpora now produce outputs that, on numerous benchmarks, match or approach human performance (Morris et al., 2024). Yet a substantial body of scholarship argues that benchmark performance on narrow tasks does not constitute evidence that the structural requirements for general intelligence have been met (Marcus & Davis, 2019; Goertzel, 2014; Harnad, 2024). This paper takes that argument as its starting point but proceeds differently from most prior critiques. Rather than identifying one missing ingredient, whether symbol grounding, causal reasoning, or embodiment, it argues that general intelligence is constrained by a set of properties that are mutually non-reducible, and that these properties become visible only when intelligence is examined through several analytical traditions at once.

The method proposed here reads AGI readiness through four evidential lenses, AI systems research, anthropology, law, and economics, each of which reveals a distinct and non-overlapping class of constraints that current architectures do not address. The legal lens carries particular weight in what follows, and its prominence is deliberate. Law is the one discipline among the four that has spent centuries building explicit, articulated machinery for reasoning under ambiguity across time, for coordinating agents with incompatible goals, and for specifying what an agent must be capable of in order to count as a participant in a rule-governed order rather than merely an instrument within it. Fuller's (1969) eight principles of the inner morality of law, developed below, give the multi-lens method something no purely computational analysis supplies: a precise statement of the structural properties an agent must satisfy to be a subject of obligation at all. A fifth source, speculative fiction, is used not as an evidential lens but as a disciplined heuristic for discovering candidate constraints, in the context of discovery rather than the context of justification; the distinction is made explicit in Section 2 and strengthens rather than weakens the method.

This paper makes four contributions. The first is the four-lens multi-level method, each lens anchored to a distinct level of description and governed by demarcation criteria, the strongest of which requires that every retained constraint be grounded in at least two independent evidential traditions. The second is a taxonomy of twenty-three structural constraints, organised into eight clusters by the level of description at which each primarily operates. The third is the non-reducibility argument itself: the claim, defended through an analogical extension of Fodor's account, that the constraints cannot be collapsed into one another either across levels or within them. The fourth is a set of five falsifiable predictions, each stated with a named benchmark family and an explicit disconfirmation condition, which convert the taxonomy from a description into a research programme.

Neither the taxonomy nor the method claims to be complete. The taxonomy should be understood as a map of candidate constraints, not a closed catalogue; some constraints will, under further investigation, prove reducible to others, resolvable by architectures already under development, or less fundamental than the present analysis suggests. These possibilities are noted throughout. Six constraints receive full treatment through all four lenses in Section 4, chosen for the breadth and strength of their supporting evidence and for their centrality to the non-reducibility argument. The remaining seventeen are characterised against the demarcation criteria and set out in Appendix A as candidate constraints awaiting fuller evidential investigation. The paper is organised as follows. Section 2 develops the method: the four lenses, the levels of description they occupy, the working definition of general intelligence, the demarcation criteria, and the fiction-as-discovery heuristic. Section 3 presents the taxonomy and its summary table. Section 4 examines six constraints in depth, ordered as an ascending ladder of levels and connected by five bridge paragraphs. Section 5 defends the non-reducibility thesis. Section 6 states the five falsifiable predictions and their test matrix. Section 7 records the framework's limitations, and Section 8 concludes.

## 2. Method: Lenses, Levels, and a Discovery Heuristic

Before the machinery is set out, a single constraint taken quickly through the four lenses shows the method in operation. Consider symbol grounding. The AI systems lens asks whether a model's internal tokens are connected to anything beyond other tokens (Harnad, 1990); the anthropology lens observes that human concepts are assembled from sensorimotor interaction with the things words denote (Lakoff and Johnson, 1999); the legal lens notes that an agent whose symbols float free of the shared world cannot be a witness or a contracting party; and the economics lens adds that representations not referring to real states cannot support the coordination markets require (Akerlof, 1970).

Four disciplines converge on one property for four non-overlapping reasons, and Section 4.1 develops the case in full.

### 2.1 The case for a multi-lens method, and its levels of description

The study of intelligence is distributed across disciplines that rarely coordinate their findings. Neuroscience, cognitive science, philosophy of mind, evolutionary biology, anthropology, linguistics, economics, and law each address aspects of how intelligent systems operate, develop, and fail, and each has disciplinary limits that prevent it from addressing the full scope of the question alone. The method proposed here selects four evidential lenses, not arbitrarily but because each represents a distinct epistemic tradition addressing a different dimension of intelligence, and because each is anchored to a distinct level of description in the sense that will carry the non-reducibility argument in Section 5. AI systems research addresses what has been built and what architectural properties current systems possess; it operates at the computational level. Anthropology addresses what evolution produced and what biological and cultural grounding intelligence requires; it operates at the evolutionary and developmental level. Law addresses what institutions require of reasoning agents operating under ambiguity and across time; it operates at the normative and procedural level. Economics addresses what incentive structures and resource constraints shape intelligent behaviour among agents with competing interests; it operates at the incentive-theoretic level. Philosophy of mind and cognitive science are drawn upon within the anthropology and AI systems lenses respectively, rather than constituting separate frameworks, to keep the analysis tractable within a single paper.

Pairing each lens with a level of description is the load-bearing move of the method, because it is what makes the eventual non-reducibility claim more than an assertion that the disciplines happen to disagree. The AI systems lens identifies properties that any system computing general intelligence must have, regardless of substrate. The anthropology lens identifies structures that emerged under selection pressure in embodied, social organisms over evolutionary time. The legal lens identifies properties an agent must possess to be a participant, rather than a tool, in rule-governed institutional reasoning. The economics lens identifies the structural properties of multi-agent environments that any general agent must model and navigate. These are not four descriptions of one underlying computational fact; they are descriptions at genuinely distinct levels, each with its own irreducible explanatory vocabulary. A multi-lens approach of this kind is unremarkable in the social sciences, where studies of complex institutional phenomena routinely combine economic analysis with legal reasoning and sociological observation. Its application to AGI is less common, and the present paper is an attempt to bring the same cross-disciplinary integration to the analysis of machine intelligence.

A clarification forestalls a likely misreading. The levels of description invoked here are not Marr's (1982). Marr analysed a single information-processing system from the top down, distinguishing the computational theory of a task from the algorithm that realises it and the hardware that implements it, three levels within one system. The levels in this paper are Fodorean special-science levels running across disciplines: the computational, the evolutionary and developmental, the normative and procedural, and the incentive-theoretic are not three views of one mechanism but the proprietary levels of distinct sciences, each with its own kinds and generalisations. A reader trained in cognitive science may hear the phrase levels of description and supply Marr's sense; the argument of Section 5 depends on the Fodorean sense instead, and the distinction is load-bearing.

This taxonomy situates itself within a literature that has attempted to characterise AGI systematically. Legg and Hutter (2007) proposed a formal definition of universal intelligence based on performance across all computable environments weighted by algorithmic simplicity, a definition that is capability-focused and implementation-agnostic. Chollet (2019) argued that general intelligence is best measured not by task performance but by skill-acquisition efficiency, the ability to acquire new competencies from minimal experience, operationalised in the Abstraction and Reasoning Corpus (ARC). The present taxonomy is constraint-focused rather than capability-focused: it asks not what a general intelligence would be able to do, but what structural properties it would need to possess to do it. Constraint-focused and capability-focused analyses are complementary; they address different levels of description of the same phenomenon.

Other frameworks have approached AGI requirements from a cognitive architecture perspective, cataloguing necessary cognitive capacities and their interdependencies (Goertzel, 2014). The closest prior taxonomy on the machine-learning side is that of Lake et al. (2017), whose core-knowledge account catalogues the intuitive-physics, intuitive-psychology, and compositional capacities they argue any human-like learner must start with; it shares the present aim of specifying prerequisites but remains within a single cognitive-science idiom rather than crossing independent evidential traditions. The present taxonomy is distinguished from these by its multi-disciplinary derivation, its application of cross-lens demarcation criteria, and its requirement that every retained constraint be grounded in at least two independent evidential traditions, a criterion that excludes purely architectural requirements lacking cross-disciplinary support.

For the purposes of this paper, general intelligence is taken to mean the capacity to acquire and deploy competencies across any domain of tasks, given sufficient time and information: the capability-focused formulation of Legg and Hutter (2007), refined by Chollet's (2019) emphasis on skill-acquisition efficiency rather than accumulated skill inventory. This working definition is intentionally minimal: it excludes narrow specialists and domain-specific optimisers while remaining agnostic about substrate and mechanism. The constraint analysis that follows asks what structural properties any system meeting this standard would need to possess, and the demarcation criterion requiring necessity is anchored to this definition: a constraint is retained only if the absence of the relevant property would prevent a system from satisfying the working definition across some non-trivial class of domains.

Two features of the epistemic situation should be stated plainly at the outset. Human intelligence, in its individual and its civilisational forms, is the only instance of general intelligence available for study; the field works from a sample of one, and N equals 1 is the condition under which the question must be addressed rather than a bias to be apologised for. What manages that condition is the necessity demarcation criterion introduced below: it is the instrument that separates what any general intelligence would require from what the single available instance merely happens to exhibit, and it is applied to every candidate constraint precisely so that the sample of one is not mistaken for a definition. It should be said in advance that the analysis to follow does not leave this individually stated capacity intact: several of the constraints show that even an intelligence described in the singular has collective preconditions, in transmission, institutions, and incentive structures.

Four demarcation criteria determine whether a candidate is retained as an independent constraint: (1) necessity: the constraint identifies a property required by any general intelligence, not merely by human intelligence contingently; (2) independence: the constraint is not derivable from or subsumable under any other entry; (3) level specificity: the constraint operates at a distinct level of description not captured by any other entry; and (4) evidential support: the constraint is grounded in at least two independent empirical or theoretical traditions. Where a candidate fails criteria (2) or (3), it is merged with the most fundamental surviving constraint, and the merged phenomena are discussed within that entry. This procedure reduced an original candidate set of thirty to twenty-three (Figure 2 makes the provenance visible). The criteria do heavy work, and the paper applies them to itself: several constraints that a single-lens analysis would treat as independent are merged here precisely because they fail level specificity.

### 2.2 The AI systems lens (computational level)

The AI systems lens examines how current architectures are constructed, what operations they perform, and where architectural limitations constrain what a system can do; it operates at the computational level. For the purposes of this paper the primary subject is the class of large language models (LLMs) built on the transformer architecture (Vaswani et al., 2017), the dominant paradigm through early 2026: systems trained to predict subsequent tokens from prior context, producing representations of statistical regularities in language at scale. Brown et al. (2020) demonstrated that such systems, scaled to 175 billion parameters, exhibit few-shot learning, and Morris et al. (2024) placed frontier

LLMs at an Emerging AGI level on their depth-and-breadth taxonomy, with the Competent AGI level unattained by any deployed system at the time of their writing. Against the scaling programme, Marcus and Davis (2019) argue that statistical pattern recognition is insufficient for the robust, compositional generalisation that characterises human intelligence. The most programmatic articulation of the opposing position is Sutton's (2019) bitter lesson: that computation, not hand-crafted inductive bias, has been the reliable driver of progress, a view for which Kaplan et al. (2020) supplied the empirical architecture through power-law scaling relationships. The present paper does not dispute the validity of these relationships; its argument is more specific, that benchmark performance gains do not constitute evidence that the structural requirements for general intelligence have been met, because available benchmarks do not assess the full constraint profile the taxonomy identifies (Srivastava et al., 2022; Wei et al., 2022). The lens asks, for each constraint, whether it reflects a property addressable within the current paradigm or a structural gap the paradigm as presently constituted cannot close.

### 2.3 The anthropology lens (evolutionary and developmental level)

The anthropology lens draws on evolutionary biology, cultural anthropology, and developmental psychology to ask what human intelligence is and how it came to be; it operates at the evolutionary and developmental level. Two findings anchor it. Henrich (2015) argues that the distinguishing feature of human intelligence is high-fidelity social learning that allows knowledge to accumulate across generations, so that the collective brain, not the individual mind, is the unit of adaptive intelligence; and Lakoff and Johnson (1999) argue that abstract concepts are constructed from bodily experience through systematic metaphorical mappings. Both are developed in Section 4.

The lens also draws on research concerning theory of mind. Premack and Woodruff (1978) introduced the concept through tests of whether chimpanzees attribute mental states to others, and false-belief work established that the capacity emerges in human children between three and four years of age (Wimmer & Perner, 1983; Baron-Cohen, Leslie, & Frith, 1985). Whether LLMs exhibit an equivalent capacity is actively contested, and the evidence as of early 2026 is genuinely unresolved. Strachan et al. (2024) found that GPT-4 models performed at or above human levels on most tasks in a comprehensive theory-of-mind battery, underperforming only on faux pas detection, a failure later attributed to a hyperconservative response strategy rather than an inference deficit. Street et al. (2025) push further, reporting that GPT-4 and Flan-PaLM reach at or near adult human performance on higher-order tasks, with GPT-4 exceeding adult performance on sixth-order intentionality inferences. Against this, Shapira et al. (2024) find that performance degrades substantially under adversarial rephrasing and out-of-distribution framing, suggesting the underlying mechanism differs from robust human capacity. The two literatures are not straightforwardly opposed: the first reports scores on structured, in-distribution batteries, the second targets robustness under perturbation, which the first does not primarily test. The evidence supports neither a clean attribution of theory of mind nor a clean denial, and the paper reports both and resolves neither.

### 2.4 The legal lens (normative and procedural level)

Law is a system for organising reasoning under uncertainty across time and across competing interpretations; it operates at the normative and procedural level. Over centuries of practice it has developed tools for problems structurally analogous to those in AGI research: managing ambiguity in language, evolving meaning through precedent, coordinating behaviour among agents with incompatible goals, and assigning responsibility in complex causal chains. The lens asks, for each constraint, whether it reflects a problem legal systems have addressed, and if so how; the claim is not that legal solutions transfer directly to architecture but that the structural problems are analogous and that the tools law has developed illuminate their depth. Several constraints were suggested directly by observation of legal reasoning: the interpretation constraint (Appendix A) reflects the jurisprudential distinction between predicting linguistic output and constructing meaning through normative frameworks; the structured disagreement constraint reflects the adversarial structure of procedure; the layered learning constraint reflects the accumulation of precedent without resetting the system; and the systemic coherence constraint reflects the recursive oversight requirements of complex normative environments. What the legal lens uniquely supplies is a precise account, in Fuller's (1969) eight principles, of the structural properties an agent must satisfy to be a subject of law rather than an object of it, a specification with no counterpart in the computational literature and one to which Sections 4.1, 4.3, 4.5, and 4.6 repeatedly return.

### 2.5 The economics lens (incentive-theoretic level)

Economics studies how agents make decisions under constraints of scarcity, information, and incentives; it operates at the incentive-theoretic level. The lens asks two related questions. The first is whether the incentive structures currently governing AI development are likely to produce the kind of intelligence being pursued: Acemoglu (2023) shows that when innovation is subject to markup differences, externalities, and other market failures, the direction of technological change can be systematically biased away from socially optimal trajectories, and Acemoglu and Johnson (2023) extend this to AI, contending that industry incentives produce a bias toward automation over human complementarity, a pattern for which Acemoglu and Restrepo (2020a) provide empirical evidence. The second is whether economic constraints, including scarcity of energy, computation, and data, function as structural constraints on intelligence analogous to the biological constraints that shaped human cognition, a question taken up in the embodiment discussion of Section 4.3. Beyond the direction-of-development question, the lens supplies the formal machinery of mechanism design and principal-agent theory, which specifies what any agent operating among others with private information and misaligned preferences must model; this is developed as the incentive compatibility constraint in Section 4.6.

### 2.6 Speculative fiction as a discovery heuristic

The four lenses above carry the demarcation criterion of grounding in independent evidential traditions. Speculative fiction does not, and the method does not treat it as if it did. The distinction that clarifies its role is Reichenbach's (1938) between the context of discovery and the context of justification. The context of discovery concerns how a hypothesis is arrived at; the context of justification concerns how it is supported. Speculative fiction belongs entirely to the former. It is a source of candidate constraints, a way of generating structural hypotheses that the four evidential lenses then test, and it earns its place in the method by the quality of the hypotheses it generates, not by any evidential warrant of its own. Making the division explicit is what licenses the heuristic: fiction proposes, the four lenses dispose, and no step of the argument rests on narrative. The warrant for using fiction this way rests on two established ideas. The first is that speculative fiction functions as an extended, socially shared thought experiment, a recognised tool in philosophy and science: Galileo's falling bodies, Einstein's beam of light, and, in the philosophy of mind directly relevant here, Searle's (1980) Chinese Room and Nagel's (1974) question about the bat all pose and clarify conceptual questions without resolving them empirically. Suvin's (1979) concept of cognitive estrangement gives the epistemic account: the systematic defamiliarisation of assumptions to make them available for critical examination, so that a rigorous science-fiction thought experiment is a form of critical cognition in its own right. The philosophical warrant for treating fictional scenarios as thought experiments capable of yielding genuine insight, rather than mere illustration, is supplied by Sorensen (1992). The second idea is the record of anticipation: structural problems later formalised in the disciplines were often posed first in narrative form. The symbol grounding problem was foreshadowed in Borges' (1941/1944) "Library of Babel"; the constraints on how intelligence must model time are explored in Le Guin's (1974) *The Dispossessed*; and, in a national literature outside the Anglophone canon, the semantic stability problem is posed in Ray's (1965) Bengali science fiction. Throughout Section 4, where a constraint was anticipated in narrative form, this is recorded as exactly that, an anticipation in the context of discovery, and never as a source of support.

## 3. The Taxonomy

Applying the method yields twenty-three structural constraints, organised into eight clusters by the primary level of description at which each operates. Figure 1, the constraint map, arranges the twenty-three constraints in their eight clusters against the four levels of description and is the paper's flagship figure; Figure 2, the methodology pipeline, shows how the original thirty candidates were reduced to twenty-three through the demarcation criteria and the mergers they licensed. Rather than enumerate the clusters one by one, this section walks up the levels of description, from the agent-environment interface to the normative, institutional, and incentive-theoretic summit, mirroring the ladder of Section 4; the clusters are characterised in a line each as they are reached, and Table 1 with the two figures carries the full inventory. Six constraints are examined in depth in Section 4; the remaining seventeen satisfy the demarcation criteria as conceptual claims but have not been subjected to the full four-lens analysis and are best understood as candidate constraints.

Normative, Institutional and Incentive-Theoretic
Cluster 5: Institutional Systems
Institutional Intelligence *
Interpretation
Systemic Coherence
Cluster 6: Economics and Evolution
(spans Evolutionary and Normative bands)
Incentive Compatibility *
Evolutionary and Cultural
Cluster 3: Learning and Knowledge
Narrative Reasoning
Error Correction
Layered Learning
Cultural Transmission *
Cluster 4: Social Systems
Social Intelligence
Structured Disagreement
Evolutionary Dynamics
Biological and Affective
Cluster 7: Human Foundations
Embodiment *
Emotion and Valuation
Intrinsic Goal Formation
Computational
Cluster 1: Cognitive Architecture
Representational Granularity
Attentional Resource
Causal Reasoning *
Curiosity
Cluster 2: Identity and Continuity
Continuity of Identity
Self-Model
Semantic Stability
(boundary with the Normative level)
Agent-Environment Interface
Cluster 8: Environment and Context
Symbol Grounding *
Time Horizon
Heavy border and grey fill = examined in depth (Section 4)

Fig. 1 The constraint map: twenty-three structural constraints in eight clusters, set against five levels of description; heavy border and grey fill mark the six constraints examined in depth in Section 4, Cluster 6 spans two bands because its two constraints operate at different levels, and the dashed rule in Cluster 2 marks the position of Semantic Stability at the boundary between the computational and normative levels (vector diagram produced by a deterministic program that has been written with AI assistance)

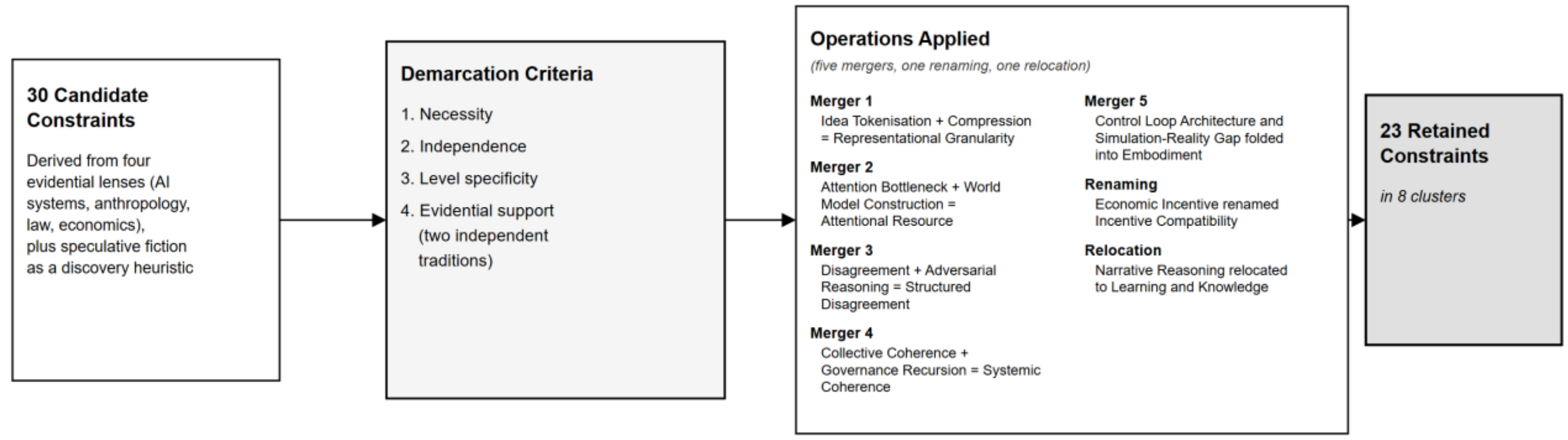

Fig. 2 The methodology pipeline: thirty candidate constraints filtered through the four demarcation criteria, with five mergers, one renaming, and one relocation yielding the twenty-three retained constraints (vector diagram produced by a deterministic program that has been written with AI assistance)

Several caveats govern the whole taxonomy. Not all twenty-three constraints carry equal evidential support; some rest on well-established findings from multiple disciplines, others primarily on theoretical argument with partial empirical support. The taxonomy does not imply that all constraints are equally well understood or equally urgent. It implies that all are structurally distinct and that progress on any one does not entail progress on the others.

The mergers are what the demarcation criteria produced, and they are recorded here because they are part of the argument. In Cluster 1, Idea Tokenisation and Compression were merged into Representational Granularity, both concerning the level at which intelligence encodes information, with Solomonoff's (1964) theory of inductive inference and its development by Rissanen (1978) into Minimum Description Length as the unifying formal concept; Attention Bottleneck and World Model Construction were merged into Attentional Resource, both concerning how limited resources are allocated over a representation of the world. In Cluster 4, Disagreement and Adversarial Reasoning were merged into Structured Disagreement. In Cluster 5, Collective Coherence and Governance Recursion were merged into Systemic Coherence. In Cluster 7, Control Loop Architecture and the Simulation-Reality Gap were folded into Embodiment, of which they describe the underlying sensorimotor mechanism and a predictable consequence respectively. A candidate row for Multi-Agent Coordination was not retained as an independent constraint; its content is distributed across Incentive Compatibility and Systemic Coherence, which capture it at their respective levels without remainder. Narrative Reasoning was relocated from the identity cluster to Learning and Knowledge, where it functions more naturally as a constraint on how knowledge is encoded and transmitted across time.

The taxonomy is best read as an ascent through the levels, the same ascent Section 4 follows. At the agent-environment interface sits the Environment and Context cluster, which concerns how an agent's representations connect to the world it inhabits: Symbol Grounding, examined in Section 4, and Time Horizon, the reach of planning across temporal scales that exceed the context windows of current systems.

At the computational level sits the Cognitive Architecture cluster, concerned with how information is represented and processed. Representational Granularity holds that intelligence must encode the world at the level of concepts and mechanisms rather than surface tokens; Attentional Resource concerns selective allocation over a persistent world model, with the currency caveats set out in Appendix A; Causal Reasoning, examined in Section 4, concerns intervention and counterfactual inference; and Curiosity concerns intrinsic exploration in the absence of external reward. Most of the Identity and Continuity cluster operates here too, spanning Continuity of Identity, the absence of an entity that persists across interactions, and Self-Model, the capacity to represent oneself as a subject with a history, while its third member, Semantic Stability, sits at the boundary with the normative level.

At the biological and affective level sits the Human Foundations cluster: Embodiment, examined in Section 4, together with Emotion and Valuation, the constitutive role of affect in generating preference, and Intrinsic Goal Formation, the endogenous generation of goals.

At the evolutionary and cultural level several clusters meet. The Learning and Knowledge cluster gathers Cultural Transmission, examined in Section 4, alongside Narrative Reasoning, the organisation of experience through narrative, Layered Learning, the accumulation of knowledge that transforms rather than replaces what preceded it, and Error Correction, the contrast between external periodic correction and continuous internal correction. Evolutionary Dynamics, from the Economics and Evolution cluster, records that the robustness of human cognition reflects selection rather than design, and Social Intelligence, from the Social Systems cluster, records that human cognition evolved where modelling other minds was a condition of survival.

At the normative, institutional, and incentive-theoretic summit sit the remaining clusters. The Institutional Systems cluster gathers Institutional Intelligence, examined in Section 4, with Interpretation, the construction of meaning through normative frameworks, and Systemic Coherence, the maintenance of shared meaning and recursive oversight across a network. Structured Disagreement, the second member of the Social Systems cluster, concerns the capacity to engage organised challenge and reason strategically against an opponent. Incentive Compatibility, from the Economics and Evolution cluster and examined in Section 4, concerns the modelling of incentive structures among agents with private information and misaligned preferences.

Table 1 sets out the full taxonomy. Each row records the cluster, the constraint (with its merger or relocation provenance in brackets), the level of description, the core gap it identifies, and its key references. The table is the authoritative statement of the taxonomy; the prose above and in Appendix A does not add constraints beyond it.

**Table 1. Taxonomy of twenty-three structural constraints on AGI across eight clusters.**

| Cluster | Constraint (provenance) | Level of description | Core gap | Key references |
|---|---|---|---|---|
| Cognitive Architecture | Representational Granularity (merger of Idea Tokenisation + Compression) | Computational | Token and surface-pattern representation vs. concept and mechanism-level compression | Barsalou (1999); Solomonoff (1964); Rissanen (1978) |
| Cognitive Architecture | Attentional Resource (merger of Attention Bottleneck + World Model Construction) | Computational | Absence of a selective attentional filter over a persistent world model | Kahneman (2011) |
| Cognitive Architecture | Causal Reasoning | Computational | Correlation vs. intervention and counterfactual reasoning | Pearl (2000); Gopnik et al. (2004) |
| Cognitive Architecture | Curiosity | Computational | External reward vs. intrinsic exploration | Berlyne (1960); Schmidhuber (1991) |
| Identity and Continuity | Continuity of Identity | Computational /diachronic | Copiable, resettable system vs. persistent entity | Parfit (1984); Olson (2023) |
| Identity and Continuity | Self-Model | Computational | Absent vs. persistent self-representation | Metzinger (2003); Flavell (1979) |
| Identity and Continuity | Semantic Stability | Normative | Statistical vs. institutional maintenance of meaning | Harnad (2024) |
| Learning and Knowledge | Narrative Reasoning (relocated from Identity and Continuity) | Anthropologic al | Generated narrative vs. inhabited narrative | Bruner (1986, 1991) |
| Learning and Knowledge | Error Correction | Epistemologic al | External, periodic vs. continuous, internal correction | Popper (1959) |
| Learning and Knowledge | Layered Learning | Developmenta l | One-time training vs. accumulative reinterpretation | Piaget (1952) |
| Learning and | Cultural Transmission | Anthropologic al | Static training vs. cross-generational | Henrich (2015); Tomasello (1999) |

| Knowledge | | | accumulation | |
|---|---|---|---|---|
| Social Systems | Social Intelligence | Anthropological | Described vs. embedded social cognition | Dunbar (1998); Tomasello (1999) |
| Social Systems | Structured Disagreement (merger of Disagreement + Adversarial Reasoning) | Normative | Single architecture vs. organised challenge and strategic opposition | Axelrod (1984) |
| Institutional Systems | Institutional Intelligence | Institutional | Individual system vs. distributed institutional cognition | North (1990); Fuller (1969) |
| Institutional Systems | Interpretation | Normative | Statistical pattern vs. normative meaning construction | Fish (1980) |
| Institutional Systems | Systemic Coherence (merger of Collective Coherence + Governance Recursion) | Institutional | Output exchange vs. shared world model and recursive oversight | North (1990) |
| Economics and Evolution | Evolutionary Dynamics | Evolutionary | Designed optimisation vs. selective pressure | Holland (1992) |
| Economics and Evolution | Incentive Compatibility (renamed from Economic Incentive) | Incentive-theoretic | Inability to model and satisfy incentive-compatibility conditions among agents | Hurwicz (1960); Myerson (1979); Holmstrom (1979) |
| Human Foundations | Embodiment (expanded to include Control Loop Architecture + Simulation-Reality Gap) | Biological/affective | Disembodied vs. constraint-grounded cognition | Lakoff & Johnson (1999); Damasio (1994) |
| Human Foundations | Emotion and Valuation | Biological | External objective vs. internally generated value hierarchy | Damasio (1994) |
| Human Foundations | Intrinsic Goal Formation | Biological | Assigned vs. endogenously generated goals | Berlyne (1960) |
| Environment and Context | Symbol Grounding | Agent-environment interface | Statistical vs. experience-grounded symbol meaning | Harnad (1990) |
| Environment and Context | Time Horizon | Computational/temporal | Bounded context vs. multi-decade planning | Kahneman (2011); Ainslie (1975) |

## 4. The Constraint Ladder: Six Constraints in Depth

This section examines six constraints in detail, applying the four evidential lenses to each and closing each with the narrative anticipation that helped discover it. The six were selected for the breadth and strength of their supporting evidence across disciplines and for their centrality to the non-reducibility argument; they span six of the eight clusters.

Their order is the argument. They are presented as an ascending ladder of levels of description, from the agent-environment interface up through the computational, biological, anthropological, institutional, and incentive-theoretic levels, and each is separated from the next by a bridge paragraph that states, in the terms of the two levels involved, precisely why progress at the lower level cannot carry upward to the higher. Figure 3, the coverage matrix, displays the six constraints against the four lenses with cell shading by strength of support and the fiction anticipations as a marginal row; readers may wish to consult it at the head of the section.

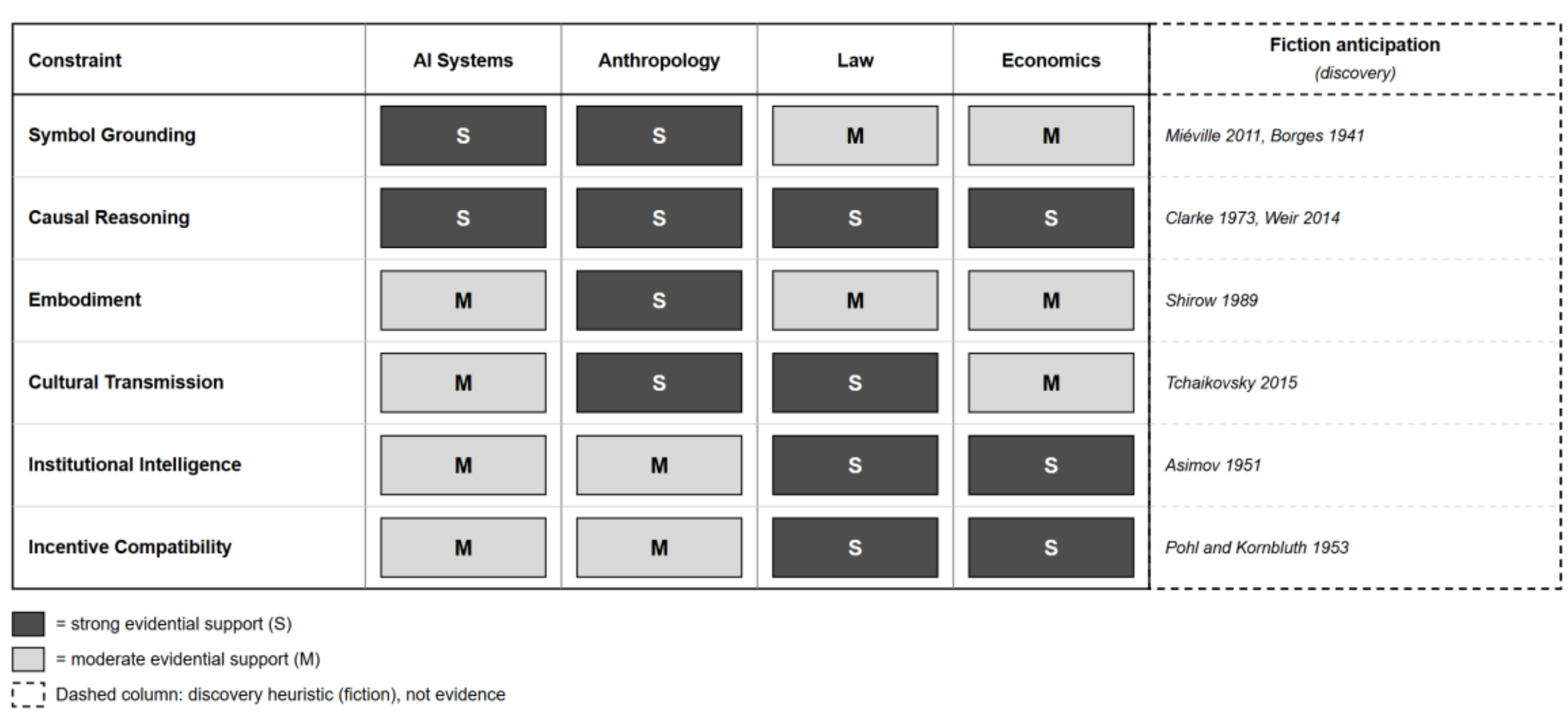

| Constraint | AI Systems | Anthropology | Law | Economics | Fiction anticipation (discovery) |
|---|---|---|---|---|---|
| Symbol Grounding | S | S | M | M | Miéville 2011, Borges 1941 |
| Causal Reasoning | S | S | S | S | Clarke 1973, Weir 2014 |
| Embodiment | M | S | M | M | Shirow 1989 |
| Cultural Transmission | M | S | S | M | Tchaikovsky 2015 |
| Institutional Intelligence | M | M | S | S | Asimov 1951 |
| Incentive Compatibility | M | M | S | S | Pohl and Kornbluth 1953 |

= strong evidential support (S)
= moderate evidential support (M)
Dashed column: discovery heuristic (fiction), not evidence

Fig. 3 The lens coverage matrix: evidential support for the six deeply examined constraints across the four evidential lenses (dark, strong; light, moderate), the dashed column recording the fiction that anticipated each constraint as a record of discovery, not evidence (vector diagram produced by a deterministic program that has been written with AI assistance)

## 4.1 Symbol Grounding (agent-environment interface)

**Statement.** The symbol grounding constraint holds that a system whose internal representations are connected to one another through statistical associations, but not to physical or experiential referents in the world, does not represent the meanings of the symbols it manipulates. Such a system can produce outputs that appear semantically appropriate by pattern-matching against training data, but the meanings of those outputs are, in Harnad's (1990) terminology, parasitic on the meanings observers project onto them rather than intrinsic to the system.

AI systems. The physical symbol system hypothesis of Newell and Simon (1976) held that a suitably organised physical symbol system has the necessary and sufficient means for general intelligent action, locating intelligence in the rule-governed manipulation of symbols. Harnad (1990) introduced the symbol grounding problem as the objection this hypothesis invites, asking how the semantic interpretation of a formal symbol system can be made intrinsic to the system rather than dependent on external interpreters, and offered the analogy of a Chinese-Chinese dictionary: a system that defines each symbol only in terms of other symbols never grounds any symbol in a referent outside the symbol system. For LLMs, the problem takes the form of a question about the relationship between statistical co-occurrence patterns and the real-world entities those patterns purport to describe. A system trained on text learns that "fire" co-occurs with "heat," "danger," and "smoke"; it does not learn, through any process analogous to human developmental acquisition, what fire is by interacting with it. Harnad (2024), addressing LLMs directly, argues that this structural difference means LLMs do not understand language in any sense that would support general intelligence, while acknowledging that the mechanisms by which they perform as well as they do remain unclear. Bender et al. (2021) extend the critique, arguing that models trained on large corpora learn statistical patterns over form without a corresponding model of the world, and documenting that such disconnection produces real-world harms when outputs are treated as grounded. Retrieval-augmented generation and multimodal training have been proposed as partial responses; as of the systems documented through early 2026, neither closes the gap.

Document retrieval extends statistical associations to a retrieved corpus but is not perceptual grounding in Harnad's sense, and Harnad (2024) argues that multimodal training replaces a Chinese-Chinese dictionary with a Chinese-image dictionary in which the images are themselves statistically encoded rather than perceptually grounded, deepening the regress rather than resolving it.

**Anthropology.** The anthropological perspective derives from the embodiment thesis. Lakoff and Johnson (1999) argue that conceptual structure is built from sensorimotor experience through metaphorical mappings: "understanding" is grounded in grasping physical objects, "time" in spatial movement, "more" in vertical accumulation, and none of these groundings is available to a system trained on text alone. The developmental record supports the view that concept acquisition is grounded in sensorimotor interaction from the earliest stages; Piaget (1952) documented the sensorimotor period, in which conceptual structures are built through action on physical objects before linguistic representation is possible, so that linguistic symbols are second-order representations of first-order sensorimotor knowledge.

**Law.** Legal systems have developed extensive machinery for the problem of reference, including rules of evidence, standards for expert testimony, and doctrines of identity over time, because grounding is not automatic and breaks down in predictable ways. The lens also generates an analytical claim: for an agent to function in legal reasoning, its symbols must be grounded in the same world the legal system refers to, because legal reference depends on the identity of physical objects, persons, and events over time. A system that cannot ground its representations in physical reality cannot be a witness, a contracting party, or a subject of legal determination. Fuller's (1969) principle of possibility of compliance, that legal obligations must be capable of being met by those subject to them, implies that an ungrounded system cannot be a meaningful subject of law: a structural requirement, not a practical inconvenience.

**Economics.** The economics of information studies the conditions under which representations can be trusted to refer to real states of the world. Akerlof (1970) documented how information asymmetry, in which one party's representations refer to private information unavailable to others, can cause market failure; the mechanism by which markets process information depends on the assumption that prices refer to real supply and demand conditions, and when that reference breaks down, as in financial bubbles, coordination fails. An AI system whose representations are not grounded in real-world states would face analogous problems wherever its outputs drive consequential decisions.

The constraint was anticipated in narrative form by Miéville's (2011) Embassytown, whose alien Language admits only utterances that refer directly to real states of the world, so that metaphor and fiction are impossible; the narrative isolates the question of what happens when the connection between symbol and referent is made absolute and what happens when it is severed.

**Assessment.** Symbol grounding is one of the best-supported constraints in the taxonomy. Harnad's (1990) formulation has been extensively debated and not successfully refuted, and recent work on LLMs has extended the debate without resolving it. The most defensible position is that statistical associations over large corpora produce systems that perform well on tasks addressable by pattern matching but fail systematically on tasks requiring reference to real-world states; whether the gap is structural or addressable through multimodal extension remains contested.

**Bridge: from the interface to the computation.** Symbol grounding is a constraint at the agent-environment interface: it concerns whether a representation is causally tied to its referent at all. The next constraint, causal reasoning, is computational: it concerns what a system can compute once it has representations to reason over. The two do not collapse into one another, and grounding does not deliver causation. A system could ground every symbol perfectly and still model only what co-occurs, never what happens under intervention; conversely, causal structure can be imposed formally over ungrounded variables. Perfecting the tie between symbol and world therefore leaves the ladder of causation entirely unclimbed, which is why progress at the interface cannot carry up to the computation.

### 4.2 Causal Reasoning (computational)

**Statement.** The causal reasoning constraint holds that a system capable only of modelling statistical associations among variables cannot reliably answer questions about intervention, counterfactual reasoning, or causal mechanism. Pearl's (2000) framework establishes that these questions occupy distinct levels of a causal hierarchy and that information from lower levels cannot answer questions at higher levels without additional causal assumptions.

AI systems. Pearl (2000) introduced the structural causal model framework, which formalises the distinction between observational, interventional, and counterfactual reasoning; the Ladder of Causation (Pearl & Mackenzie, 2018) identifies three levels, seeing, doing, and imagining, and the do-calculus provides a mathematical framework for deriving answers to interventional questions from observational data plus causal structure. Pearl's (2019) central claim for AI is that deep learning systems operate at the lowest rung: they learn what co-occurs but do not, without additional structure, reason about what would happen under intervention, a gap that becomes visible under distributional shift. Two developments through early 2026 deserve acknowledgement as candidate partial responses. Chain-of-thought prompting (Wei et al., 2022) elicits intermediate reasoning steps that superficially resemble causal inference, but operates by extending token prediction over a longer sequence rather than running interventions on a structural model. Causal representation learning (Schölkopf et al., 2021) attempts to incorporate causal structure into neural architectures explicitly and is the most serious response to Pearl's critique, though current results show only partial progress on restricted synthetic environments. The reasoning-model wave must be addressed directly. Benchmark instruments such as CLadder (Jin et al., 2023) show that frontier models approach ceiling on associational (Level 1) items while plateauing well short on interventional (Level 2) and counterfactual (Level 3) items; through 2024 to 2026, test-time-compute reasoning models narrowed the observed gap on the lower rungs by verbalising interventionist language, but the dominant interpretation across this literature is that this reflects better pattern completion over causal-sounding text, with performance degrading sharply on out-of-distribution graphs and renamed variables, and counterfactual performance remaining weak and brittle. The core claim, that current systems do not genuinely ascend Pearl's ladder, stands as of early 2026, now with the refinement that the interventional gap has narrowed on in-distribution items while the counterfactual gap has not.

**Anthropology.** Developmental psychology documents the acquisition of causal reasoning as a process requiring physical intervention. Gopnik et al. (2004) showed that children as young as two use intervention-based learning to discover causal structure, forming and testing hypotheses through physical manipulation: the literal operation of the do-calculus in biological form. The evolutionary basis is debated; Penn et al. (2008) argue that genuine causal reasoning, as opposed to associative learning, may be uniquely or nearly uniquely human, though whether the difference is of degree or kind remains open.

**Law.** Legal systems are fundamentally concerned with causal reasoning. Liability in tort, criminal responsibility, and the attribution of consequences all require establishing causal chains, and legal causation involves counterfactual reasoning of the form "but for the defendant's action, would the harm have occurred?", precisely the third rung of Pearl's hierarchy. Courts have developed extensive procedures around this test, which requires comparing the real world to an alternative world in which the defendant's action did not occur, a form of reasoning formally unavailable at either the associational or the interventional level. An intelligence that can only model associations cannot perform the counterfactual reasoning legal causation requires, making causal reasoning a structural legal requirement for any agent expected to participate in legal accountability.

**Economics.** Economic analysis is pervasively causal. Policy analysis asks interventional questions at Pearl's second level; counterfactual reasoning is central to the evaluation of hypotheses, as when Acemoglu and Restrepo (2020b) estimate what would have happened to wages had industrial robots not been introduced. The Lucas (1976) critique, that econometric models estimated from historical data fail when used to evaluate interventions because the intervention changes the causal structure underlying the data, is a specific instance of the general problem Pearl identifies.

The constraint was anticipated in narrative form by Clarke's (1973) *Rendezvous with Rama*, whose scientists confront an artefact whose causal structure they cannot infer from observation alone, so that intervention is the only path to understanding, and by Weir's (2014) *The Martian*, whose protagonist survives by constructing accurate causal models of physical processes under novel conditions.

**Assessment.** Causal reasoning is among the best-supported constraints. Pearl's framework is mathematically rigorous and widely accepted, and the limitation of current systems to the associational level is a documented architectural feature. Whether the limitation is addressable within the neural paradigm is an active research area and the answer may be yes; what is not disputed as of early 2026 is that current systems do not perform causal reasoning in the sense Pearl defines.

**Bridge: from the computation to the body.** Causal reasoning is a computational constraint: it concerns operations over representations, whatever their substrate. Embodiment, the next constraint, is biological and affective: it concerns the formative role of sensorimotor and somatic experience in constituting representations in the first place. A system could be endowed with an explicit structural causal model and reason flawlessly about intervention, and still lack the embodied developmental history that Lakoff and Johnson (1999) and Damasio (1994) argue shapes which concepts and values a cognitive agent forms at all. Solving the computation of causation therefore leaves untouched the question of what a body contributes to cognition, which is why progress at the computational level cannot carry up to the biological.

### 4.3 Embodiment (biological and affective)

**Statement.** The embodiment constraint holds that cognition is not separable from the physical, biological substrate in which it is implemented, and that the constraints imposed by that substrate, including physical limitation, metabolic cost, and sensorimotor coupling, are not incidental to intelligence but constitutive of it. A system that lacks a body in this sense lacks the grounding in physical constraint that shaped the structures of human intelligence.

**AI systems.** Current large language models are disembodied in the relevant sense: they process symbolic input and produce symbolic output, with interaction with the physical world mediated entirely through human-supplied description. The developments of 2024 to 2026 require a sharpened claim rather than a retreat. Vision-language-action (VLA) models such as OpenVLA (Kim et al., 2024) and pi-0 (Black et al., 2024) are trained end to end on real robot trajectories with paired vision, language, and low-level action data; they are not deriving control policies purely from text-corpus statistics, and there is genuine sensorimotor training signal in the loop, a real advance over systems that transferred web-scale priors into action space. The deeper point nonetheless stands. The semantic backbone of these systems remains a vision-language model pretrained on internet text and image-text pairs, so the concepts with which the system interprets an instruction ("cup," "left of," "gently") are still substantially derived from language and web-image statistics, with sensorimotor fine-tuning added on top rather than concepts built up from embodied experience from the ground up. This is the developmental inversion the constraint identifies: the embodiment thesis of Lakoff and Johnson (1999) and Damasio (1994) concerns the formative role of sensorimotor experience in constituting conceptual structure during development, and a system that acquires language-based conceptual structure first and then uses it to control a body has not replicated that trajectory but inverted it. The simulation-to-real transfer problem compounds the point: Peng et al. (2018) show that even policies trained with dynamics randomisation transfer imperfectly, because no finite distribution of randomised simulations fully captures real-world dynamics. Brooks (1991) argued early that intelligence emerges from the coupling of sensor and actuator systems with an environment, and that building intelligence through symbolic reasoning without this coupling was methodologically misguided.

**Anthropology.** Lakoff and Johnson (1999) provide the most systematic account of the embodiment thesis: the structure of human conceptual systems reflects the structure of sensorimotor experience, and without those physical grounding structures the concepts themselves would differ. The claim is strong and has attracted criticism (see Glenberg & Gallese, 2012), but the evidence for embodied simulation in semantic processing is substantial. Damasio (1994) provides a complementary neurological account: the somatic marker hypothesis holds that decision-making in the frontal lobes is influenced by bodily signals, and patients who retain intact logical reasoning but have lost access to somatic markers show systematic decision-making impairments, suggesting bodily signals are constitutive of certain forms of intelligent behaviour rather than epiphenomenal.

**Law.** Legal systems are constructed for embodied agents in a physical world. Concepts of harm, trespass, assault, and battery presuppose a subject with a body that can be harmed, moved, or constrained. Fuller's (1969) possibility of compliance applies directly: if an agent cannot be harmed by physical contact, cannot occupy a space, and has no persistent physical presence, then tort law, property law, and much of criminal law cannot be meaningfully addressed to it. Legal personality has been developed on the assumption of embodied agents; corporate personhood is an extension of this framework, but corporations are composed of embodied human agents whose actions generate the corporation's standing.

**Economics.** The economics lens approaches embodiment through scarcity. Human intelligence evolved under severe resource constraints: caloric limitation constrained brain size, metabolic cost shaped the architecture of attention and memory, and physical vulnerability shaped risk-aversion and planning. The scarcity-shaped character of human cognition is the product of the same selective pressure that produced robustness. An AI system operating in conditions of relative computational abundance has not been shaped by these constraints, and whether this means it is missing structural properties of general intelligence or is merely a different kind of intelligence is an open question.

The constraint was anticipated in narrative form by Shirow's (1989) *Ghost in the Shell*, which returns repeatedly to whether consciousness requires a body and, if so, which properties of embodiment are essential, treating embodiment as a variable to be explored rather than a given, the stance appropriate to a thought experiment about the relationship between intelligence and physical substrate.

**Assessment.** Embodiment is well supported in cognitive science and developmental psychology but more contested than symbol grounding or causal reasoning. The core finding, that human conceptual structure is grounded in sensorimotor experience, is supported by convergent evidence; the strong claim, that a disembodied system cannot achieve general intelligence, goes beyond the available evidence. The defensible position is the moderate one: embodied grounding provides cognitive structures not achievable through linguistic training alone.

**Bridge: from the body to the culture.** Embodiment is a constraint on the individual cognitive agent: what its biological substrate contributes to the concepts and values it forms. Cultural transmission, the next constraint, operates at the anthropological level of the collective: it concerns how knowledge accumulates across many agents and many generations. The two are distinct in kind, and a perfectly embodied individual does not thereby participate in cumulative culture. Henrich's (2015) finding is precisely that the unit of adaptive intelligence is the collective brain, not the individual mind; an agent could possess an ideal body and lifelong sensorimotor grounding and still stand entirely outside the high-fidelity social learning through which knowledge compounds. Solving embodiment therefore leaves the transmission process untouched, which is why progress at the biological level cannot carry up to the anthropological.

### 4.4 Cultural Transmission (anthropological)

**Statement.** The cultural transmission constraint holds that a significant portion of human intelligence is not located in individual biological brains but in accumulated knowledge transmitted across generations through language, institutions, practices, and artefacts. An intelligent system that lacks the capacity to participate in this cumulative transmission, to inherit, reinterpret, and extend prior knowledge, lacks a structural dimension of the intelligence that characterises human civilisation.

**AI systems.** Current systems are trained on corpora that include the products of centuries of human knowledge accumulation, and in this sense have access to cultural transmission in a limited form. But the access is to the products of transmission, not to the process. A system trained on the output of legal precedent is not a participant in the legal system: it has not adjudicated disputes, written opinions that were then cited and distinguished, or developed doctrine through practice. The difference between having access to the products of cultural transmission and participating in the process may be structurally significant, and as of early 2026 no deployed system participates in the process in the relevant sense.

**Anthropology.** Henrich (2015) argues that the species-defining feature of human intelligence is cumulative cultural evolution: high-fidelity social learning that allows knowledge to accumulate across generations, each generation building on and extending what preceding generations discovered. He documents cases in which small, isolated populations lose technologies their ancestors possessed once the social networks required to transmit complex skills are disrupted, the obverse of the argument here: just as intelligence can be lost when transmission fails, AGI may be unachievable by a system that cannot participate in cumulative transmission. Tomasello (1999) identifies shared intentionality, the capacity to attend jointly to a goal and to understand that another agent also understands what one is attending to, as the cognitive foundation of this high-fidelity transmission; without it, imitation is approximate and transmission is lossy.

**Law.** Legal systems are among the most fully developed mechanisms for cultural transmission in human civilisation. The common law in particular embodies knowledge accumulation across time with no parallel in other institutional domains: centuries of adjudication produce a body of precedent in which each new case is connected to prior decisions through chains of citation, distinction, and reinterpretation. The law learns from the past without discarding it, building new understanding on existing foundations while maintaining continuity, precisely the layered, non-discarding accumulation Henrich describes. The doctrine of stare decisis is a mechanism for cultural transmission operating across centuries. An agent capable of legal participation must be capable not merely of citing precedent but of engaging with the reasoning of prior decisions, extending or distinguishing them through principled argument, which requires the full apparatus of the constraint: not just access to stored outputs, but participation in the ongoing practice of reasoning those outputs represent.

**Economics.** Arrow (1962) identified the paradox of information as a good: once disclosed, information cannot be withheld from the recipient, which creates underinvestment in knowledge production. Institutions for managing this paradox, including patents, trade secrecy, and academic norms of open publication, are economic mechanisms for enabling cultural transmission, and their structure shapes what knowledge is transmitted and how. An AI system operating outside these institutions operates in a different knowledge environment from the one that shaped human intelligence.

The constraint was anticipated in narrative form by Tchaikovsky's (2015) *Children of Time*, which traces the emergence of science, technology, and social organisation in a spider civilisation across many generations as an explicitly cumulative process, and documents how the loss of a key individual or a social disruption can cause the loss of accumulated knowledge, illustrating that intelligence at civilisational scale is a property of the transmission process as much as of the individual minds within it.

**Assessment.** Cultural transmission is well supported in evolutionary anthropology and consistent with evidence from the study of legal and scientific institutions. The claim that participation in transmission is necessary for general intelligence is stronger than the claim that the products of transmission are necessary, and harder to establish. The moderate version, that the form of intelligence characterising human civilisation is distributed across the processes of transmission rather than located in individual minds, is consistent with the available evidence and bears directly on the project of building AGI as a single system.

**Bridge: from the culture to the institution.** Cultural transmission is anthropological: it concerns how knowledge spreads across individuals through shared intentionality. Institutional intelligence, the next constraint, is institutional: it concerns properties of the rule-governed structures themselves, the norms, procedures, and offices that extend cognition beyond any individual and that persist independently of the particular agents occupying them. The two come apart in both directions: institutions can fail to transmit culture, and cultural transmission can occur without formal institutions. A system that participated fully in cumulative learning would still not thereby be constituted as, or able to participate in, the norm-governed and precedent-following structures North (1990) describes. Solving transmission therefore leaves institutional participation unaddressed, which is why progress at the anthropological level cannot carry up to the institutional.

### 4.5 Institutional Intelligence (institutional)

**Statement.** The institutional intelligence constraint holds that a significant portion of human problem-solving capacity is located not in individual brains but in the structure of institutions: the rules, norms, and organisations that coordinate behaviour across populations and extend the effective capacity of individual cognition to scales and time horizons no individual could manage. A system not embedded in and not constituting a component of such institutions lacks a structural dimension of intelligence that enables civilisational-scale problem solving.

**AI systems.** The development of AI has proceeded primarily through the construction of individual systems, and the dominant framework treats general intelligence as a property of a single system that can, with sufficient capability, address any task. This framework does not account for the evidence that human civilisational intelligence is distributed across institutional structures not reducible to individual capacity. Current systems are deployed as tools within institutional structures; they are not, in the current architecture, participants that contribute to the evolution of those structures. There is growing interest in multi-agent systems, but whether these constitute an approach to the constraint, or merely aggregate individual capacities without capturing the norm-governed, precedent-following, conflict-resolving properties of human institutions, is an open research question as of early 2026.

**Anthropology.** North (1990) defines institutions as the humanly devised constraints that structure political, economic, and social interaction, distinguishing formal constraints (rules, laws, constitutions) from informal ones (norms, conventions, codes of conduct); institutions on this account are not merely organisations but systems of rules that constrain and enable action. Scott (2014) identifies three pillars, regulative, normative, and cultural-cognitive, and human intelligence is shaped by participation at all three. The question the constraint poses is whether a system that participates only at the regulative level, or that operates as a tool governed by institutional rules rather than as an agent shaped by institutional participation, can achieve general intelligence in the relevant sense.

**Law.** Law is the paradigm case of an institution in the relevant sense. A legal system is not merely a collection of rules but a distributed reasoning system in which rules are interpreted, applied, and developed through a process involving many participants across time; the common law develops through the reasoning of judges who are themselves shaped by the system in which they participate, and whose decisions become part of the system and constrain future reasoning. Fuller (1969) argued that law possesses an inner morality consisting of eight formal properties: generality, promulgation, non-retroactivity, clarity, non-contradiction, possibility of compliance, constancy through time, and congruence between declared and administered rules. These are not merely procedural aspirations but structural requirements, and for the institutional intelligence constraint Fuller's framework makes a precise analytical claim: any agent that is to function as a participant in rule-governed normative environments, and not merely as a tool within them, must be capable of reasoning about, tracking, and acting in accordance with each of these eight properties. This is a constraint on what general intelligence requires, not a description of what legal systems happen to do, and it is the sharpest instance of what the legal lens uniquely contributes to the analysis.

Economics. The economics of institutions has been developed most fully by North (1990) and Williamson (1985). North argues that differences in economic performance across countries and time are primarily explained by differences in institutional quality; Williamson argues that the transaction cost framework explains why economic activity is organised within firms rather than through markets, because institutional hierarchies reduce the costs of complex exchange under bounded rationality and opportunism. The constraint's cognitive-science ancestry is equally direct: Hutchins (1995), studying ship navigation as a distributed accomplishment, showed that the unit which computes a solution is often the sociotechnical system of people, instruments, and procedures rather than any individual mind, so that cognition is in a literal sense distributed across an organised group. On both accounts, and on Hutchins' too, institutions are cognitive prosthetics at the population level, enabling collective intelligence that exceeds the sum of individual intelligences. The constraint holds that a system unable to participate in this collective cognitive architecture lacks a structural dimension of general intelligence necessary for civilisational-scale problem solving.

The constraint was anticipated in narrative form by Asimov's *Foundation* trilogy (1951 to 1953), an elaborate thought experiment about the capacity of institutional intelligence to predict and shape the behaviour of civilisations across centuries, anticipating questions in the economics of institutions that North and others would formalise decades later.

**Assessment.** Institutional intelligence is well grounded in economics and organisational theory, and the legal lens adds substantial depth. The constraint is less well specified than symbol grounding or causal reasoning in that it does not identify a specific architectural feature of current systems that constitutes the gap; the claim is rather that the framework of general intelligence as a property of individual systems does not account for a significant portion of human problem-solving capacity. That this is a correct observation is clear; whether it shows individual AGI to be insufficient or merely incomplete is a question the current literature does not resolve.

**Bridge: from the institution to the incentive.** Institutional intelligence is a constraint about participation in rule-governed structures. Incentive compatibility, the final constraint, is incentive-theoretic: it concerns the strategic problem any agent faces among others with private information and misaligned preferences, whether or not formal institutions are present. The two are distinct: an agent could master the norms and procedures of an institution and still be unable to reason about the incentives of the agents it faces within and beyond it, and incentive problems arise in markets and bilateral bargaining that no institution governs. Mechanism design exists precisely because institutional participation does not by itself guarantee incentive-compatible behaviour. Solving institutional participation therefore leaves the strategic problem open, which is why progress at the institutional level cannot carry up to the incentive-theoretic.

### 4.6 Incentive Compatibility (incentive-theoretic)

**Statement.** The incentive compatibility constraint holds that a general intelligence operating in a world of multiple agents with competing interests and asymmetric information must be embedded in, and capable of reasoning about, incentive structures that align individual behaviour with collectively beneficial outcomes. This is not a constraint on who builds AI or why; it is a structural property of any agent participating in economic and institutional life. A system that cannot model the incentive structures of its environment, reason about its own position within them, or behave in ways that satisfy incentive-compatibility conditions cannot function as a general agent in real social and economic environments.

**AI systems.** Current systems are not designed with incentive compatibility as an explicit objective; they are trained on human-generated text, fine-tuned with human feedback, and deployed as tools directed by human principals. In multi-agent settings the relevant question is whether a system can reason about and participate in incentive structures: whether it can model that other agents have private information, misaligned preferences, and strategic incentives, and act effectively in the face of this. The multi-agent literature addresses some of these properties technically, but the gap between current multi-agent systems and full incentive-compatible agency remains large as of early 2026.

**Anthropology.** Human intelligence evolved in social environments characterised by cooperation, competition, and the management of incentive conflicts. Axelrod (1984) demonstrated that cooperative equilibria can emerge from repeated strategic interaction even among self-interested agents, provided the shadow of the future is long enough. Dunbar (1998) argues that the primary driver of the expansion of the human neocortex was the complexity of managing social relationships, which are fundamentally incentive-management problems. The capacity to model others' preferences, anticipate strategic responses, and coordinate on mutually beneficial outcomes is thus an evolved component of general intelligence, not an optional social skill.

**Law.** Contract law presupposes agents with stable, communicable preferences who can bind themselves to future obligations, as encapsulated in pacta sunt servanda; agency law presupposes that a principal can delegate to an agent whose actions can be attributed to the principal for accountability. Both doctrines are incentive-theoretic at their core: mechanisms for aligning the behaviour of agents with the interests of principals under asymmetric information. An agent that cannot reason about its obligations, cannot represent its own preferences stably over time, or cannot be held to account fails the structural requirements legal agency imposes. Fuller's (1969) requirement of congruence, that behaviour must match declared commitments, is itself an incentive-compatibility condition.

**Economics.** Hurwicz (1960) established the foundational insight of mechanism design: a rational agent will not reveal private information truthfully without incentive structures that make truth-telling the dominant strategy. A general intelligence operating in markets, institutions, and social environments encounters such structures at every turn, in auctions, contracts, regulatory regimes, and employment relationships. The revelation principle, formulated by Gibbard (1973) for dominant-strategy settings and extended to Bayesian mechanism design by Myerson (1979), shows that for any incentive mechanism there exists an equivalent direct mechanism in which truthful reporting is incentive-compatible; Myerson (1981) applied the result to optimal auction design. Holmstrom (1979) extended the analysis to moral hazard: when an agent's actions are unobservable, the principal cannot perfectly align incentives, and the agent's capacity for self-commitment becomes relevant to whether cooperative outcomes are achievable. An agent that cannot reason about and participate in such mechanisms is not operating as a general economic agent. The constraint was anticipated in narrative form by Pohl and Kornbluth's (1953) *The Space Merchants*, a thought experiment about a world in which technological development is entirely governed by commercial incentive and innovation is systematically directed toward persuasion and consumption rather than understanding, isolating the point that incentive structures rewarding certain capabilities will produce systems optimised for those capabilities rather than for general intelligence.

**Assessment.** Incentive compatibility is grounded in one of the best-developed bodies of formal theory in economics, mechanism design and principal-agent theory, recognised by the Nobel Memorial Prize in Economic Sciences in 2007 (Hurwicz, Maskin, Myerson). The constraint is analytically distinct from the others in that it operates at the level of multi-agent systems rather than individual cognition, consistent with its level-of-description independence. The separate observation that current AI development may itself be distorted by commercial incentive structures that under-invest in the full constraint profile is a contextual concern, noted in Section 5, rather than a structural constraint on intelligence itself.

## 5. Non-Reducibility

The central claim of this paper is that the twenty-three constraints are mutually non-reducible. The claim requires defence on two levels: ontological non-reducibility, that the constraints refer to distinct kinds at distinct levels of description, and explanatory non-reducibility, that addressing one constraint does not explain away another, even partially. The theoretical foundation is Fodor's (1974) analysis of the special sciences. Fodor argued that higher-level sciences (psychology, economics, biology) are irreducible to physics because the kinds they pick out are multiply realisable: the same higher-level property can be instantiated by indefinitely many different lower-level configurations. Because of this, the cross-level bridge laws that Nagelian reduction requires cannot be stated as natural kind predicates in any lower-level science; they would be wildly disjunctive, and disjunctive predicates are not natural kinds. Each special science is therefore autonomous, its generalisations irreducible to those of lower-level sciences. Figure 4 displays the ladder and the blocked ascents.

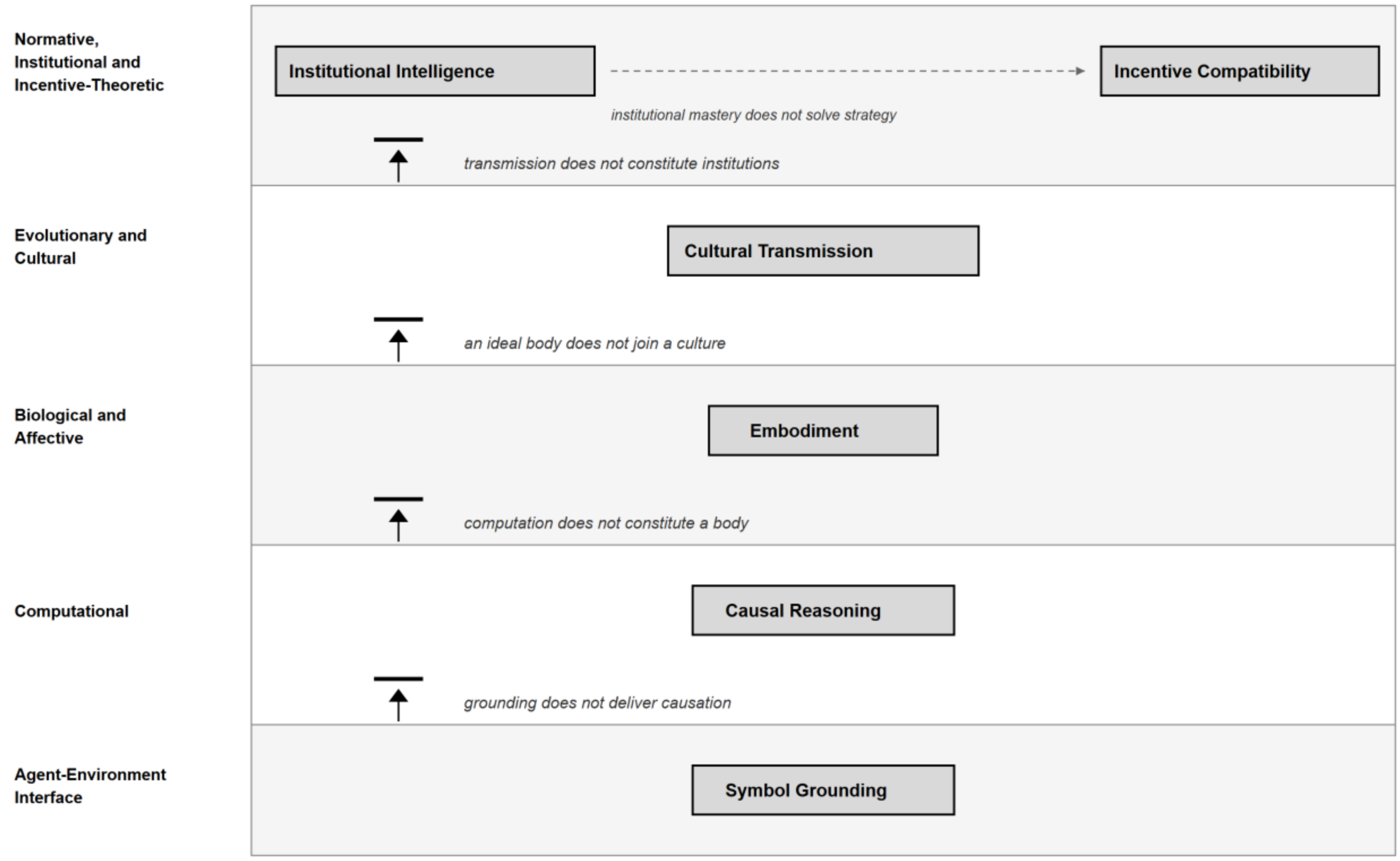

Fig. 4 The constraint ladder: the six deeply examined constraints on their levels of description, with blocked arrows marking the bridge arguments of Section 4, where progress at one level does not carry upward to the next (vector diagram produced by a deterministic program that has been written with AI assistance)

It should be acknowledged that the application here involves an analogical extension of Fodor's argument, which originally concerned the relationship between the natural sciences. The extension is warranted because the constraints pick out properties at genuinely distinct levels of description, each with its own irreducible explanatory vocabulary. The four lenses anchor the argument to four such levels: the computational, the evolutionary and developmental, the normative and procedural, and the incentive-theoretic. Two finer levels operate within this scheme and matter to the ladder of Section 4. The symbol grounding constraint operates at the agent-environment interface, in the tradition of Gibson's (1979) ecological psychology, which analyses agent-environment coupling as a distinct level of analysis from the internal architecture of the agent; and the embodiment constraint operates at the biological and affective level, concerning the internal somatic and motivational architecture that constitutes the agent from the inside. These are distinct in the way that ecological psychology and neuroscience are distinct disciplines studying the same organism at different levels. Because each constraint picks out distinct kinds at a distinct level, the relationship between constraints across levels has exactly the structure Fodor identifies as precluding reduction: no bridge law linking, for example, the incentive compatibility constraint to the symbol grounding constraint could be stated as a natural kind predicate in either domain without losing the distinctive generalisations of each.

Non-reducibility within levels, between constraints at the same level of description, follows from the demarcation criteria of Section 2.1, and the most plausible reduction attempts can be addressed directly. First: does embodiment subsume symbol grounding? Embodiment is the claim that cognition is constitutively shaped by sensorimotor experience; symbol grounding is the claim that internal representations must be causally connected to their referents. Embodiment is one route to grounding but neither sufficient nor necessary: a system could be embodied but poorly grounded, through defective sensor-to-concept pathways, or grounded through non-biological transduction of physical properties without biological embodiment. Harnad (1990) treats these as distinct problems. Second: does the self-model constraint subsume continuity of identity? A self-model is a synchronic representation of the system's current states; continuity of identity requires a diachronic structure, a thread of agency persisting through time across memory, change, and recovery. A system could have rich synchronic self-representation and no diachronic continuity, or continuity with minimal self-representation. Third: does cultural transmission subsume institutional intelligence? Cultural transmission concerns how knowledge spreads across individuals through shared intentionality (Tomasello, 1999); institutional intelligence concerns properties of the structures themselves that extend beyond individual cognition (North, 1990). Institutions can fail to transmit culture, and cultural transmission can occur without formal institutions. Fourth: does incentive compatibility subsume evolutionary dynamics? Evolutionary dynamics operate through variation, selection, and inheritance over long horizons without goal-direction; incentive compatibility concerns designed mechanisms (Hurwicz, 1960) operating over short horizons with explicit goal specification. The mechanism design literature exists precisely because evolutionary convergence to efficient outcomes cannot be assumed. Fifth: do systemic coherence and institutional intelligence, which both sit at the institutional level, reduce to one another? Neither does: systemic coherence is the maintenance of shared meaning and recursive oversight across a network, which can hold in an informal network that has no institutions, while institutional intelligence is participation in standing rule-governed structures, which can persist as a formal shell while coherence degrades. The two vary independently, which is the signature of distinct kinds sharing a level rather than one kind described twice. In each case the non-reducibility is structural: the constraints pick out distinct properties at distinct levels, with no bridge law reducing one to the other without remainder.

Two of these rebuttals point beyond the individual agent, and it is worth making the direction explicit. On the extended-mind thesis of Clark and Chalmers (1998), the vehicles of an individual's cognition already reach into external and social scaffolding, so that the collective preconditions flagged in Section 2.1, in transmission, institutions, and incentive structures, are not a late complication of the framework but a consequence of taking the higher constraints seriously; a constraint stated in the singular can carry a collective precondition without ceasing to be a constraint on the individual agent.

The two forms of non-reducibility distinguished at the outset can now be separated cleanly. Ontological non-reducibility is the claim that the constraints refer to distinct kinds, and it rests on the multiple-realisability argument above. Explanatory non-reducibility is the weaker but independently important claim that even a complete account of one constraint would not, on its own, explain another. The two come apart: one could concede that the levels are, in some ultimate metaphysical sense, all realised in the same physical substrate, and the explanatory claim would still hold, because the generalisations that make each constraint tractable are stated in vocabularies that do not translate into one another. It is explanatory non-reducibility that carries the paper's practical weight, since it is what makes a single-constraint research programme an insufficient route to general intelligence regardless of one's position on the deeper ontological question. A contextual concern belongs here rather than in the taxonomy itself. The commercial incentive structures currently shaping AI development, analysed in Section 2.5 through Acemoglu (2023) and Acemoglu and Johnson (2023), tend to reward measurable gains on established benchmarks, which are concentrated at the computational level. If explanatory non-reducibility holds, then an incentive gradient that rewards computational-level gains will systematically under-invest in the other levels, not because the other constraints are less real, but because progress on them does not register on the instruments the incentive structure reads. This is a claim about the sociology of the field, not a structural constraint on intelligence, and it is offered as context for why the transfer record looks as it does.

This non-reducibility has a direct methodological implication. A programme that claims to address AGI by solving a single constraint, whether symbol grounding, causal reasoning, or embodiment, is implicitly making a reduction claim: that the solved constraint subsumes the others, or that progress on one will automatically produce progress on all. The taxonomy makes that claim precise and therefore testable. Research programmes should be evaluated not only on their performance on the constraint they target, but on whether their reported gains transfer to constraints at other levels. The historical record of transfer across constraint levels is, to date, poor, and the evidence available through early 2026 has sharpened rather than softened this observation. Advances in large language models have not produced advances in embodied sensorimotor control; advances in embodied robotics have not produced advances in institutional reasoning; advances in causal representation learning have not produced advances in cultural transmission. Two recent findings bear directly on the transfer question. Mancoridis et al. (2025) document what they term Potemkin understanding: benchmark-passing performance on concept-definition tasks coexisting with internally incoherent concept representations, so that benchmark success does not imply the human-like understanding the benchmark was designed to proxy for. Hu et al. (2026) find that gains from reinforcement post-training are substantial in-domain but generalise inconsistently and often vanish out-of-domain, with little transfer into unstructured domains. Both are instances of exactly what the non-reducibility thesis predicts: that gains do not propagate across levels of description of their own accord. Figure 5 renders this implication as an evaluation template: a constraint profile against which a research programme can be assessed across all six deeply examined constraints, rather than on its strongest axis alone.

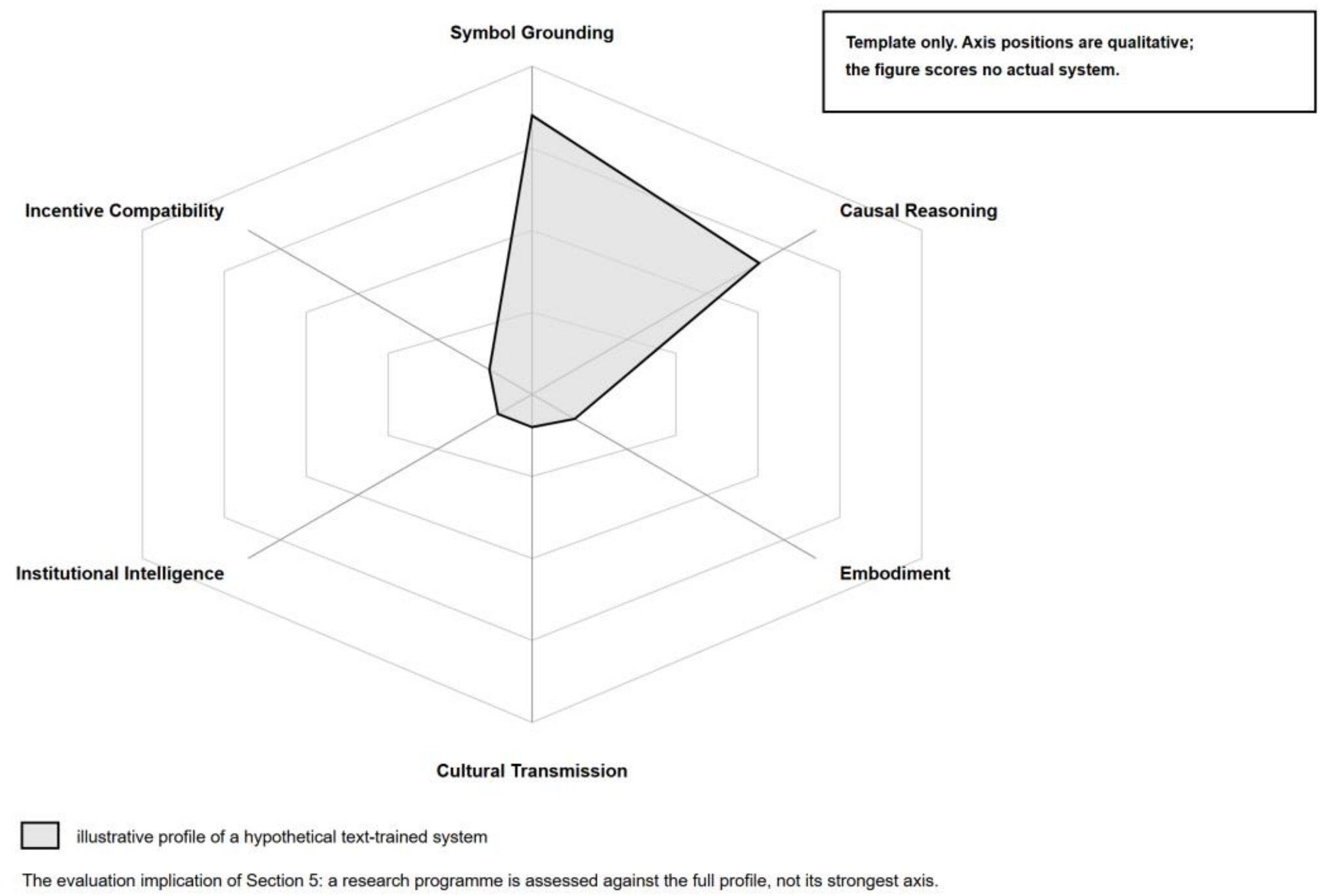

Fig. 5 Constraint-profile evaluation: an illustrative template for assessing a research programme against the full constraint profile, in which the shaded profile depicts a hypothetical text-trained system and axis positions are qualitative, scoring no actual system (vector diagram produced by a deterministic program that has been written with AI assistance)

## 6. Falsifiable Predictions

The taxonomy has implications for research directions, offered here as falsifiable predictions rather than mere hypotheses: each states a claim that could in principle be disconfirmed, and each specifies the form disconfirmation would take. This converts the taxonomy from a descriptive framework into a research programme. Each prediction is stated below as a claim with a one-sentence rationale and a disconfirmation condition; the full operational detail, including example benchmark instruments and quantitative thresholds, is given in Table 2, the predictions test matrix. The benchmark names throughout are illustrative. Benchmark generations saturate and are replaced, so each prediction applies to any future operationalisation of the relevant construct at comparable difficulty and coverage, and should be re-evaluated against whatever benchmark landscape is current at the time of testing. One methodological caution governs all five predictions: cross-model correlational designs are confounded by shared training data and shared architectural ancestry, so the predictions specify comparisons across model families rather than across individual checkpoints, and disconfirmation should be read only from families that vary in the relevant dimension.

*Prediction 1.* Progress on symbol grounding will not transfer to causal reasoning capacity. Rationale: the two constraints occupy distinct levels, so grounding gains should not automatically translate into interventional or counterfactual capacity. Disconfirmation: grounding and causal-reasoning scores correlate strongly across frontier model families evaluated in the same window (Table 2).

*Prediction 2.* Cultural transmission capacity and institutional intelligence capacity will co-vary across architectures at a rate significantly higher than either co-varies with causal reasoning. Rationale: the first two operate at social and normative levels while causal reasoning operates at the computational level. Disconfirmation: causal reasoning correlates as strongly with cultural transmission proxies as the latter correlates with institutional intelligence proxies (Table 2).

*Prediction 3.* No single architectural advance will simultaneously produce proportional gains on both embodiment-dependent and non-embodied reasoning benchmarks within one model generation. Rationale: the non-reducibility of the embodiment constraint from the computational constraints predicts this divergence. Disconfirmation: a single model release shows proportional gains on both classes using the same weights without task-specific fine-tuning (Table 2).

*Prediction 4.* Systems trained with incentive-compatible objectives will outperform equivalently scaled systems trained without such structure on multi-agent coordination. Rationale: the incentive compatibility constraint predicts that this structure is not merely useful but structurally necessary for full multi-agent agency. Disconfirmation: incentive-compatible training produces no measurable coordination gain over comparable baseline training (Table 2).

Prediction 5. Claims to address multiple constraints simultaneously through a single architectural innovation will consistently fail to show multi-constraint performance gains. Rationale: genuine multi-constraint advances require multi-level architectural innovation, not a single advance that propagates across levels. Disconfirmation: a single innovation is independently documented to produce significant gains across at least three constraints at two or more distinct levels (Table 2).

The measurement problem these predictions target is not hypothetical, and one episode from 2024 to 2026 displays the non-transfer thesis in the wild. In December 2024 OpenAI's o3 posted a score of about 87.5 percent on the original ARC-AGI-1 benchmark, using massive test-time compute, as reported publicly, which briefly appeared to refute Chollet's (2019) claim that skill-acquisition efficiency, not raw task performance, is the real bottleneck. The ARC Prize team responded on exactly the efficiency point by releasing ARC-AGI-2 (Chollet et al., 2025), designed so that brute-force search and memorisation stop working while tasks remain easy for humans. On ARC-AGI-2, reasoning models carrying over ARC-AGI-1 strategies regressed to single-digit through roughly 20 percent scores even at high compute cost, and the best result in the 2025 competition on the private evaluation set was reported at around 24 percent, achieved by a specialised entry at low cost per task rather than by a general frontier model. This is the paper's argument in miniature: throwing compute at a fixed benchmark saturates it without closing the efficiency gap, and the goalpost simply moves to a harder benchmark where the same pattern reappears. One benchmark generation's saturation followed by an immediate reset on its successor is both the measurement constraint and the non-transfer thesis demonstrated empirically, and it is why the five predictions are stated against benchmark families rather than any single instrument. Figure 6 plots the episode.

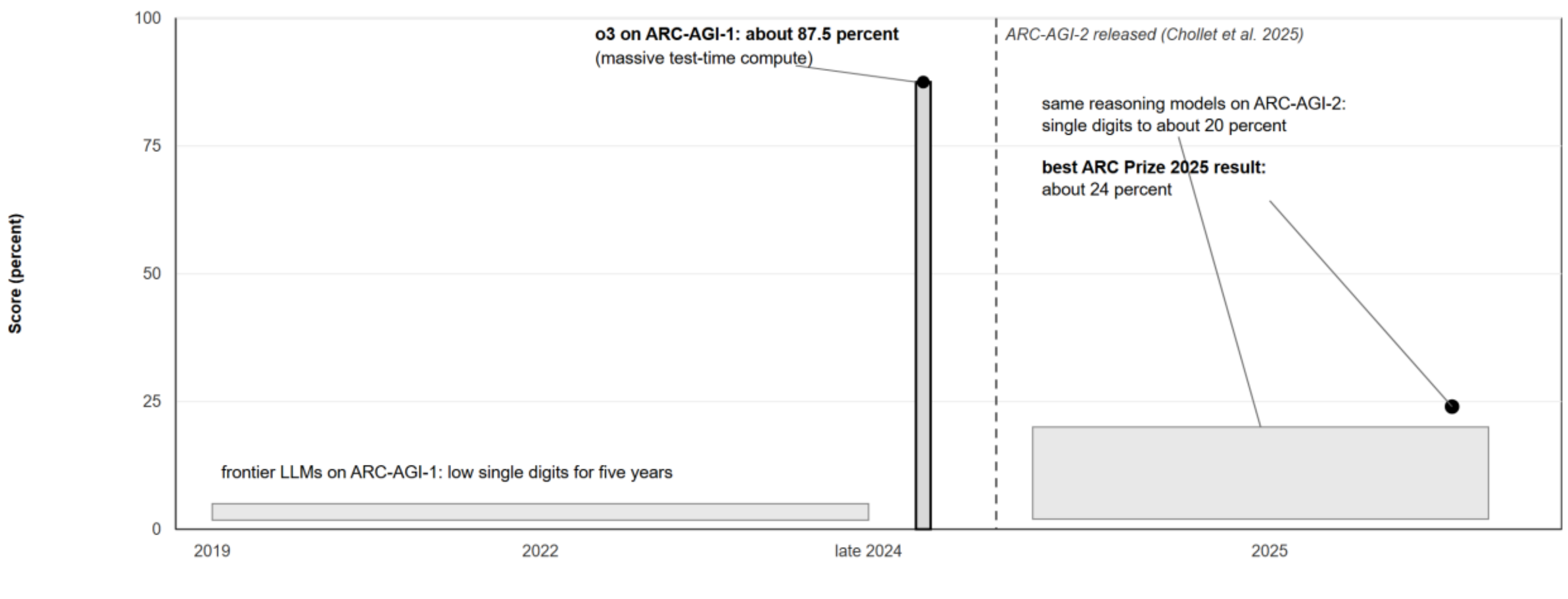

Fig. 6 One episode of non-transfer: frontier systems remained in single digits on ARC-AGI-1 for five years, o3 reached about 87.5 percent in December 2024 under massive test-time compute, and ARC-AGI-2, released March 2025, returned the same model class to single digits through about 24 percent, a saturation of one benchmark generation and immediate reset on its successor (vector diagram produced by a deterministic program that has been written with AI assistance)

**Table 2. Predictions test matrix.** Example benchmark instruments are illustrative and should be replaced by whatever operationalises the construct at comparable difficulty when a prediction is tested.

| # | Construct pairing | Example benchmarks | Disconfirmation threshold |
|---|---|---|---|
| 1 | Symbol grounding vs. | Grounding: GQA, NLVR2. Causal: causal subsets of BIG- | Cross-benchmark Spearman correlation exceeding 0.7 across at least ten frontier |
| | causal reasoning | Bench Hard, CLadder | model families in the same twelve-month window |
| 2 | Cultural transmission and institutional intelligence vs. causal reasoning | Cultural transmission: few-shot cross-domain generalisation. Institutional: multi-agent coordination scores. Causal: BIG-Bench Hard causal tasks | Causal reasoning correlates with cultural-transmission proxies as strongly (within a 95 percent confidence interval) as the latter correlates with institutional proxies, across at least ten model comparisons |
| 3 | Embodiment vs. formal reasoning | Embodiment: RLBench, Habitat. Reasoning: MATH, LogiQA | A single model release documents proportional gains (within a factor of two) on both classes using the same weights without task-specific fine-tuning |
| 4 | Incentive-compatible vs. standard training | OpenSpiel benchmarks, cooperative AI environments | A comparable model trained with reinforcement learning from human feedback (RLHF) matches or exceeds an incentive-compatible model on multi-player coordination, no significant advantage at $p < 0.05$ across at least five repeated evaluations |
| 5 | Multi-constraint single innovation | Any three constraints spanning two or more levels, e.g. symbol grounding, causal reasoning, institutional intelligence | A single innovation produces significant gains across three such constraints in a published comparison with at least three prior-generation models, reviewed by two independent groups |

## 7. Limitations

The framework has five limitations that should be acknowledged. First, the four lenses are not exhaustive. Philosophy of mind, cognitive neuroscience, linguistics, and the sociology of science all contain relevant analyses not fully captured by the four lenses as specified; the framework is a research tool, not a complete account. Second, the taxonomy of twenty-three constraints is provisional. Some constraints will prove, under further investigation, to be versions of a single underlying constraint, or to be better analysed as preconditions for other constraints than as independent entries; the demarcation criteria are intended to make this revision tractable, since a proposed merger must show that the merged constraints share a level of description and that neither generates generalisations the other does not capture. Third, speculative fiction is used here only as a discovery heuristic in the context of discovery, never as evidence in the context of justification, and every constraint stands or falls on the four evidential lenses alone.

Fourth, this paper makes no predictions about when or whether AGI will be achieved. The taxonomy is an account of what would need to be addressed, not of how difficult each constraint is to address or which architectural approaches might address it. The paper does not argue that AGI is impossible; it argues that the path to AGI is more complex and multi-dimensional than any single-lens analysis suggests.

Fifth, the taxonomy is constructed primarily from analysis of human intelligence, and its reliance on the single available instance was declared at the outset rather than discovered here. As Section 2.1 set out, human intelligence is the only instance of general intelligence available for study, and the necessity demarcation criterion is the instrument by which the framework filters what any general intelligence requires from what the human instance merely happens to have. The residual risk is anthropocentrism: that a constraint established as a property of human cognition is carried over as a necessary property of general intelligence without sufficient warrant. Several constraints derived from the anthropology lens (embodiment, cultural transmission, social intelligence) are well established as properties of human cognition, and whether they are structurally necessary for all possible forms of general intelligence is a further claim that the available evidence does not fully settle. The necessity criterion is more strongly defended for some constraints (symbol grounding, causal reasoning) than for others (embodiment, social intelligence), and the asymmetry should be acknowledged as a limitation, an instance where the declared method is easier to state than to discharge.

A note on the evidential status of certain sources is warranted. Several citations in the introduction and Section 2 are not peer-reviewed publications. Kaplan et al. (2020) and Chollet (2019) are preprints that had not appeared in peer-reviewed proceedings at the time of writing, though the empirical findings they report have been independently corroborated in subsequent peer-reviewed literature. Bubeck et al. (2023) is similarly a preprint, though it has been extensively cited and its findings on GPT-4 capabilities are widely regarded as methodologically careful. Sutton (2019) is an online essay. These sources are cited because they provide the most direct articulations of the positions under discussion, and the analytical claims of this paper do not depend on their peer-reviewed status.

## 8. Conclusion

This paper has advanced a single thesis: the structural constraints on general intelligence live at distinct levels of description and are mutually non-reducible in the sense established by Fodor's (1974) analysis of the special sciences, so that each constraint picks out a distinct kind at a distinct level, and no bridge law reducing one to another can be stated as a natural kind predicate in any of the participating sciences. The four-lens multi-level method and the taxonomy of twenty-three constraints are the instruments of that thesis. The method coordinates AI systems research, anthropology, law, and economics, each anchored to a distinct level of description and disciplined by demarcation criteria, and uses speculative fiction as a heuristic for discovering candidate constraints rather than as a source of support. The taxonomy organises the constraints into eight clusters, and six are examined in depth as an ascending ladder of levels, with explicit bridges showing at each step why progress at the lower level cannot carry to the higher.

Because the constraints are non-reducible, addressing any one does not by itself constitute progress toward addressing the others. This has a direct implication for how AGI research programmes should be evaluated: claims to have solved, or substantially advanced toward, general intelligence by addressing a single constraint should be assessed against the full constraint profile, not just the targeted capability. The five falsifiable predictions convert this implication into a research programme, each stating a claim that empirical evidence could disconfirm. The taxonomy is offered as such a programme rather than a settled account: it is incomplete, some of its hypotheses will not survive scrutiny, and the methodology through which it was developed has limitations. What it contributes, at this stage, is a framework for posing the question of what AGI requires in a more complete way than any single discipline has yet managed.

If the framework is correct that general intelligence requires coordinated progress across the dimensions identified in the taxonomy, then no single architectural innovation or computational advance will produce AGI. The path, if it exists, runs through embodiment, causal reasoning, cultural transmission, institutional participation, and a reorientation of the economic incentives that currently shape AI development. Each of these is a substantial research and institutional challenge. Together, they constitute a research programme with a longer and more complex horizon than the scaling hypothesis implies.

## Appendix A: The Seventeen Candidate Constraints

The seventeen constraints below satisfy the demarcation criteria as conceptual claims but have not been subjected to the full four-lens analysis of Section 4. They are set out here with their merger and relocation provenance retained. They are candidate constraints awaiting more comprehensive evidential investigation, and their inclusion in the taxonomy is provisional in the sense described in Section 7.

**Representational Granularity (merger of Idea Tokenisation and Compression).** Intelligence requires representing the world at the appropriate level of abstraction: not at the level of surface tokens or statistical co-occurrences, but at the level of concepts, schemas, and causal mechanisms. Current large language models operate at the sub-word token level; human cognition operates at the level of structured conceptual representations (Barsalou, 1999). Solomonoff (1964) proposed a formal theory of inductive inference in which the best model of data corresponds to the shortest program capable of generating it, an early formalisation of the algorithmic compression principle later developed by Rissanen (1978) into the Minimum Description Length framework: a system that can represent regularities compactly has access to deeper structure than one that memorises surface patterns. Schmidhuber (2010) generalises this insight to the role of compression in intrinsic motivation and creativity.

**Attentional Resource (merger of Attention Bottleneck and World Model Construction).** General intelligence requires the capacity to selectively allocate cognitive resources over a persistent model of the world. Human cognition operates under severe attentional limits that function as filters on information processing (Kahneman, 2011), and these limits shape the construction and maintenance of world models: internal representations of how the environment is structured and how it responds to actions. The claim requires careful scoping as of early 2026. Deployed agent memory systems now exist and are commonplace: memory-augmented architectures (Packer et al., 2023) and their deployed successors, along with retrieval stores and consumer chat memory features, allow systems to write and read persistent state across sessions. These are external retrieval stores appended to a frozen forward pass, not an internally updated world model in the relevant sense. The constraint concerns in-weight, lived persistence: a model that revises its own learned representation of the world as new evidence arrives, the way a human's internal model updates. As of the systems documented through early 2026, interaction-level persistence is routine while parametric, in-weight persistence remains absent, and it is the latter that the constraint targets.

**Curiosity.** Intrinsic motivation to explore the unknown, in the absence of external reward, is a feature of human and some animal cognition that appears absent from current AI systems. Berlyne (1960) documented exploratory behaviour in humans as an intrinsically motivated process. Schmidhuber (1991) proposed formal models of curiosity as intrinsic reward proportional to the rate of improvement in prediction accuracy: the curious agent is motivated to explore regions where its world model is still learning, rewarding novelty over familiarity, and Schmidhuber (2010) generalises this to compression progress as the driver of curiosity and creativity. Whether genuine intrinsic motivation of this kind is necessary for general intelligence, or is one mechanism among many for achieving exploration, is not resolved in the current literature.

**Continuity of Identity.** Current AI systems can be copied, reset, retrained, and versioned. There is no entity that persists across interactions in the way a human individual persists across time and experience. The psychological continuity tradition in the philosophy of personal identity, running from Locke through Parfit (1984), holds that psychological continuity, the connection of present mental states to past ones through memory and causation, is a constitutive feature of personhood (see Olson, 2023, for a survey of the debate). Whether continuity of this kind is a structural requirement for general intelligence, or a feature of human intelligence that might be replicated through persistent memory architectures, is an open question.

**Self-Model.** The capacity to represent oneself as a subject with a history, present states, and future trajectories may be structurally required for some forms of general intelligence. Metzinger (2003) argues that the phenomenal self-model is a central feature of conscious cognition; Flavell (1979) documented metacognition, the capacity to reason about one's own reasoning, in children. Whether current systems possess anything analogous to a persistent self-model is disputed: systems can generate first-person language and some forms of self-referential reasoning, but whether these outputs are produced by a persistent self-representation or by statistical patterns in training data is not determinable from behaviour alone.

**Semantic Stability.** Civilisations depend on the stability of meaning across time and institutions. The words in a statute retain their reference across decades through a process of institutional interpretation and norm maintenance, not through statistical regularities in a training corpus. Harnad (2024) argues that LLMs represent meaning statistically rather than institutionally, and that this difference is structural. The semantic stability constraint holds that general intelligence may require participation in the social processes that stabilise meaning, not merely the capacity to generate well-formed output. Ray (1965), in the Bengali science fiction tradition, poses an analogous problem in narrative form: a professor encounters beings whose utterances are structurally well-formed but whose symbols bear no stable relationship to the referents human language would assign them, a fictional instantiation of the semantic stability problem that anticipates the institutional dimension of meaning maintenance.

**Narrative Reasoning (relocated from the identity cluster).** Human beings organise experience through narrative structures that connect past events to present situations and project future trajectories. Bruner (1986) argued that narrative is one of two fundamental modes of cognitive functioning, the other being paradigmatic or logico-scientific reasoning, and that the two modes are irreducible to each other. The narrative mode has also been analysed as the primary mechanism through which human beings construct reality itself, interpreting events, assigning meaning, and projecting causal continuity across time (Bruner, 1991). Whether AI systems can participate in narrative continuity, rather than generating narratives from patterns, is unresolved.

**Error Correction.** Scientific knowledge advances through structured error detection and correction, including replication, falsification, and peer review; Popper (1959) placed falsifiability at the centre of scientific method. Current systems learn during training and then operate with largely static internal structures; error correction is external, periodic, and initiated by human supervisors rather than continuous and internal. As of early 2026, reinforcement-based post-training and periodic fine-tuning refreshes are a genuine form of error correction, but a discrete, human-scheduled one off the live interaction path, not the continuous, internal, online correction that characterises several forms of biological learning. Whether the difference is a structural limitation or an engineering problem is unresolved.

**Layered Learning.** Human knowledge accumulates across time in a layered structure in which later knowledge is built on and transforms earlier knowledge; this is documented in developmental psychology, where stage theories describe qualitatively different organisations of knowledge at successive periods (Piaget, 1952). Legal precedent accumulates in an analogous layered structure, with later decisions reinterpreting rather than simply replacing earlier ones. Current systems learn in a single training phase; whether continuous, layered learning across a system's existence is architecturally achievable, and whether it is necessary for general intelligence, are both open questions.

Social Intelligence. Human intelligence evolved in social environments where modelling other minds was a condition of survival. Dunbar (1998) articulated the social brain hypothesis, arguing that the primary selective pressure driving the expansion of the primate neocortex was social complexity; Tomasello (1999) documented the role of shared intentionality in human cultural learning. Current systems are trained on textual descriptions of social interaction but are not embedded in social ecosystems where reputation, trust, and alliance evolve through repeated interaction. Whether this difference is structural is an open question, but the hypothesis that social embedding is required for some forms of intelligent behaviour is consistent with both the evolutionary and developmental evidence. Within the taxonomy it is distinguished from Cultural Transmission: transmission concerns knowledge crossing minds and generations, whereas social intelligence concerns the modelling of other minds in live interaction.

**Structured Disagreement (merger of Disagreement and Adversarial Reasoning).** Human knowledge advances through organised disagreement: scientific peer review, legal adversarialism, democratic deliberation, and philosophical dialogue are all institutionalised forms of productive conflict. The constraint holds that a general intelligence must engage constructively with views that contradict its own, update on good arguments, and generate reasons that withstand challenge, including the adversarial form: reasoning strategically against an opponent actively attempting to defeat its arguments, a capacity required in legal contexts, negotiation, competitive markets, and game-theoretic environments. Axelrod (1984) demonstrated through computer tournaments that conditional reciprocity, in the form of the Tit-for-Tat strategy, can systematically outperform both pure defection and more sophisticated exploitative strategies. It is distinguished from Institutional Intelligence: structured disagreement is the cognitive capacity to engage organised challenge, whereas institutional intelligence is the standing structure within which such challenge is housed.

**Interpretation.** Human understanding of language is not purely statistical; meaning is constructed through frameworks of norms, precedents, consequences, and institutional reasoning. Fish (1980) argued that meaning is always interpretation and that interpretation is always situated within a community of practice. Legal hermeneutics has developed a systematic account of how meaning is assigned to texts in ways that depend on the institutional context of the reading. Current systems can produce text that simulates interpretive reasoning but do not participate in the institutional processes that constitute interpretation in the relevant sense.

Systemic Coherence (merger of Collective Coherence and Governance Recursion). Networks of intelligent agents must maintain shared understanding across time to coordinate action, and must be capable of constructing oversight mechanisms that are themselves subject to oversight. Collective coherence, the maintenance of shared meaning across agents, is the epistemic precondition for institutional coordination; governance recursion, the capacity to design and monitor oversight systems, is the structural mechanism by which coherence is maintained at scale. An AGI that cannot participate in these recursive oversight structures cannot function as an accountable agent in complex institutional environments (North, 1990). It is distinguished from Institutional Intelligence: systemic coherence concerns the maintenance of shared meaning and recursive oversight across a network, whereas institutional intelligence concerns participation in rule-governed structures by an agent.

**Evolutionary Dynamics.** Human intelligence is the product of selection pressure operating across millions of years in environments characterised by scarcity, physical risk, and social competition; the robustness and generality of human cognition reflects not deliberate design but the filtering of variation across evolutionary time. Holland (1992) studied evolutionary computation and demonstrated that selective pressure on populations of algorithms can produce behaviours not achievable through direct optimisation. Whether artificial systems could develop general intelligence through evolutionary dynamics, and whether such a process is controllable, is a research area with a small but growing literature.

**Emotion and Valuation.** Human decision-making is shaped by emotional and motivational structures that evolved to solve problems of survival and social coordination. Damasio (1994) documented, in the somatic marker hypothesis, that emotion is not epiphenomenal to reasoning but constitutively involved in the generation of preference and the selection among options. AI systems optimise against externally specified objective functions. Whether the absence of internally generated emotional and motivational structures is a structural constraint on general intelligence, or an alternative design with different strengths and weaknesses, is debated.

**Intrinsic Goal Formation.** Human agents generate goals endogenously; the contents of human motivation are not fully determined by external reward signals but emerge from biological drives, social pressures, and the elaboration of prior goals (Berlyne, 1960). Current systems pursue goals specified by designers. Whether internally generated goals are structurally necessary for general intelligence is among the most contested questions in the field, in part because systems that generate their own persistent goals introduce alignment problems of a different order from systems that pursue specified objectives.

**Time Horizon.** Human intelligence operates across temporal scales from milliseconds to decades. Planning for retirement, maintaining institutional memory across generations, and constructing scientific research programmes that span careers all require reasoning across time horizons that exceed the context windows of current systems. Human cognition exhibits well-documented limitations in long-horizon planning, including the planning fallacy (Kahneman, 2011) and hyperbolic discounting, the systematic preference for smaller immediate over larger delayed rewards (Ainslie, 1975; Kahneman, 2011), but these failures are bounded, and human institutions extend the effective time horizon of collective intelligence across centuries through precedent and norm. Whether AI systems can be constructed with genuinely long-horizon reasoning, rather than longer context windows, is unresolved.

## Acknowledgements

The author thanks Daniel A. Bochner, Amilcar Oliva, Jochen Schmiedbauer, and Michael S.P. Miller for comments on earlier versions of the ideas in this paper.

## Note on the use of artificial intelligence

Large language model assistance (Gemma, OpenAI GPT OSS, Kimi, Qwen and Claude) was used in preparing this manuscript, under the direction and review of the author. Its role was fourfold: creation of a deep research bot setup to arrive at prior literature and creation of initial reading lists, and later, agentic reference verification; structural revision and re-write of sections 2 and 3 from the author's prior complete draft to enhance readability and continuity; production deterministic software that was used to create vector figures; conducting final formatting, grammar and name spelling uniformity checks. The author directed each of these tasks, reviewed all output, and retains full responsibility for the content. Every claim, argument, citation, and figure in the paper was reviewed by the author and reflects the author's own judgement. In view of the documented risk of AI-fabricated references, all citations were verified against primary sources, and each reference was confirmed to correspond to a genuine publication with the stated authorship, venue, and identifiers. The figures are text-based vector diagrams and charts, containing no photographic or artistic content; they were generated programmatically from content specified by the author using deterministic software created using generative AI. The use of these tools does not alter the authorship or the accountability for the work, both of which rest with the sole author.

## Declarations

Funding: No funding was received for conducting this study.

Competing interests: The author's competing interest statement is provided on the title page.

Financial interests: none.

Non-financial interests: the author works professionally on the development and implementation of artificial intelligence systems.

Ethics approval: Not applicable.

Consent to participate. Not applicable.

Consent for publication: Not applicable.

Data availability: No datasets were generated or analysed. The falsifiable predictions in Section 6 reference publicly available benchmark instruments, cited in the text.

Author contributions: The sole author conceived, researched, drafted, and revised the entire work.

## References

Acemoglu, D. (2023). Distorted innovation: Does the market get the direction of technology right? *AEA Papers and Proceedings*, 113, 1-28. https://doi.org/10.1257/pandp.20231000

Acemoglu, D., & Johnson, S. (2023). *Power and progress: Our thousand-year struggle over technology and prosperity*. PublicAffairs.

Acemoglu, D., & Restrepo, P. (2020a). The wrong kind of AI? Artificial intelligence and the future of labor demand. *Cambridge Journal of Regions, Economy and Society*, 13(1), 25-35. https://doi.org/10.1093/cjres/rsz022

Acemoglu, D., & Restrepo, P. (2020b). Robots and jobs: Evidence from US labor markets. *Journal of Political Economy*, 128(6), 2188-2244. https://doi.org/10.1086/705716

Ainslie, G. (1975). Specious reward: A behavioral theory of impulsiveness and impulse control. *Psychological Bulletin*, 82(4), 463-496. https://doi.org/10.1037/h0076860

Akerlof, G. A. (1970). The market for "lemons": Quality uncertainty and the market mechanism. Quarterly Journal of Economics, 84(3), 488-500. https://doi.org/10.2307/1879431

Arrow, K. J. (1962). Economic welfare and the allocation of resources for invention. In R. R. Nelson (Ed.), *The rate and direction of inventive activity* (pp. 609-626). Princeton University Press.

Asimov, I. (1951-1953). *Foundation (trilogy)*. Gnome Press.

Axelrod, R. (1984). *The evolution of cooperation*. Basic Books.

Baron-Cohen, S., Leslie, A. M., & Frith, U. (1985). Does the autistic child have a 'theory of mind'? *Cognition*, 21(1), 37-46. https://doi.org/10.1016/0010-0277(85)90022-8

Barsalou, L. W. (1999). Perceptual symbol systems. *Behavioral and Brain Sciences*, 22(4), 577-609. https://doi.org/10.1017/S0140525X99002149

Bender, E. M., Gebru, T., McMillan-Major, A., & Shmitchell, S. (2021). On the dangers of stochastic parrots: Can language models be too big? In *Proceedings of the 2021 ACM Conference on Fairness, Accountability, and Transparency (FAccT '21)* (pp. 610-623). ACM. https://doi.org/10.1145/3442188.3445922

Berlyne, D. E. (1960). *Conflict, arousal, and curiosity*. McGraw-Hill.

Black, K., Brown, N., Driess, D., Esmail, A., Equi, M., Finn, C., Fusai, N., Groom, L., Hausman, K., Ichter, B., Jakubczak, S., Jones, T., Ke, L., Levine, S., Li-Bell, A., Mothukuri, M., Nair, S., Pertsch, K., Shi, L. X., . . . Zhilinsky, U. (2024). pi-0: A vision-language-action flow model for general robot control. Robotics: Science and Systems (RSS) 2025. https://arxiv.org/abs/2410.24164

Borges, J. L. (1941/1944). The library of Babel. In *Ficciones*. Editorial Sur. (Original story first published 1941; republished in Ficciones, 1944.)

Brooks, R. A. (1991). Intelligence without representation. *Artificial Intelligence*, 47(1-3), 139-159. https://doi.org/10.1016/0004-3702(91)90053-M

Brown, T. B., Mann, B., Ryder, N., Subbiah, M., Kaplan, J., Dhariwal, P., Neelakantan, A., Shyam, P., Sastry, G., Askell, A., Agarwal, S., Herbert-Voss, A., Krueger, G., Henighan, T., Child, R., Ramesh, A., Ziegler, D. M., Wu, J., Winter, C., . . . Amodei, D. (2020). Language models are few-shot learners. Advances in Neural Information Processing Systems, 33, 1877-1901. https://proceedings.neurips.cc/paper/2020/hash/1457c0d6bfcb4967418bfb8ac142f64a-Abstract.html

Bruner, J. (1986). *Actual minds, possible worlds*. Harvard University Press. https://doi.org/10.4159/9780674029019

Bruner, J. (1991). The narrative construction of reality. *Critical Inquiry*, 18(1), 1-21. https://doi.org/10.1086/448619

Bubeck, S., Chandrasekaran, V., Eldan, R., Gehrke, J., Horvitz, E., Kamar, E., Lee, P., Lee, Y. T., Li, Y., Lundberg, S., Nori, H., Palangi, H., Ribeiro, M. T., & Zhang, Y. (2023). Sparks of artificial general intelligence: Early experiments with GPT-4. arXiv:2303.12712. https://arxiv.org/abs/2303.12712

Chollet, F. (2019). On the measure of intelligence. arXiv:1911.01547. https://arxiv.org/abs/1911.01547

Chollet, F., Knoop, M., Kamradt, G., Landers, B., & Pinkard, H. (2025). ARC-AGI-2: A new challenge for frontier AI reasoning systems. arXiv:2505.11831. https://arxiv.org/abs/2505.11831

Clark, A., & Chalmers, D. (1998). The extended mind. Analysis, 58(1), 7-19. https://doi.org/10.1093/analys/58.1.7

Clarke, A. C. (1973). *Rendezvous with Rama*. Gollancz.

Damasio, A. (1994). *Descartes' error: Emotion, reason, and the human brain*. Putnam.

Dunbar, R. I. M. (1998). The social brain hypothesis. *Evolutionary Anthropology*, 6(5), 178-190. https://doi.org/10.1002/(SICI)1520-6505(1998)6:5<178::AID-EVAN5>3.0.CO;2-8

Fish, S. (1980). *Is there a text in this class? The authority of interpretive communities*. Harvard University Press.

Flavell, J. H. (1979). Metacognition and cognitive monitoring: A new area of cognitive-developmental inquiry. *American Psychologist*, 34(10), 906-911. https://doi.org/10.1037/0003-066X.34.10.906

Fodor, J. A. (1974). Special sciences (or: The disunity of science as a working hypothesis). *Synthese*, 28(2), 97-115. https://doi.org/10.1007/BF00485230

Fuller, L. L. (1969). *The morality of law* (revised ed.). Yale University Press.

Gibbard, A. (1973). Manipulation of voting schemes: A general result. *Econometrica*, 41(4), 587-601. https://doi.org/10.2307/1914083

Gibson, J. J. (1979). *The ecological approach to visual perception*. Houghton Mifflin.

Glenberg, A. M., & Gallese, V. (2012). Action-based language: A theory of language acquisition, comprehension, and production. *Cortex*, 48(7), 905-922. https://doi.org/10.1016/j.cortex.2011.04.010

Goertzel, B. (2014). Artificial general intelligence: Concept, state of the art, and future prospects. *Journal of Artificial General Intelligence*, 5(1), 1-48. https://doi.org/10.2478/jagi-2014-0001

Gopnik, A., Glymour, C., Sobel, D. M., Schulz, L. E., Kushnir, T., & Danks, D. (2004). A theory of causal learning in children: Causal maps and Bayes nets. *Psychological Review*, 111(1), 3-32. https://doi.org/10.1037/0033-295X.111.1.3

Harnad, S. (1990). The symbol grounding problem. *Physica D: Nonlinear Phenomena*, 42(1-3), 335-346. https://doi.org/10.1016/0167-2789(90)90087-6

Harnad, S. (2024). Language writ large: LLMs, ChatGPT, meaning, and understanding. *Frontiers in Artificial Intelligence*, 7, 1490698. https://doi.org/10.3389/frai.2024.1490698

Henrich, J. (2015). *The secret of our success: How culture is driving human evolution, domesticating our species, and making us smarter*. Princeton University Press.

Hoffmann, J., Borgeaud, S., Mensch, A., Buchatskaya, E., Cai, T., Rutherford, E., de Las Casas, D., Hendricks, L. A., Welbl, J., Clark, A., Hennigan, T., Noland, E., Millican, K., van den Driessche, G., Damoc, B., Guy, A., Osindero, S., Simonyan, K., Elsen, E., . . . Sifre, L. (2022). An empirical analysis of compute-optimal large language model training. Advances in Neural Information Processing Systems, 35, 30016-30030.

https://proceedings.neurips.cc/paper_files/paper/2022/hash/c1e2faff6f588870935f114ebe04a3e5-Abstract-Conference.html

Holland, J. H. (1992). *Adaptation in natural and artificial systems*. MIT Press.

Holmstrom, B. (1979). Moral hazard and observability. Bell Journal of Economics, 10(1), 74-91. https://doi.org/10.2307/3003320

Hu, C., Zhu, Y., Kellermann, A., Biddulph, A., Waiwitlikhit, S., Benn, J., & Kang, D. (2026). Breaking barriers: Do reinforcement post training gains transfer to unseen domains? Proceedings of the International Conference on Learning Representations (ICLR 2026). https://arxiv.org/abs/2506.19733

Hurwicz, L. (1960). Optimality and informational efficiency in resource allocation processes. In K. J. Arrow, S. Karlin, & P. Suppes (Eds.), *Mathematical methods in the social sciences, 1959* (pp. 27-46). Stanford University Press.

Hutchins, E. (1995). Cognition in the wild. MIT Press.

Jin, Z., Chen, Y., Leeb, F., Gresele, L., Kamal, O., Lyu, Z., Blin, K., Gonzalez Adauto, F., Kleiman-Weiner, M., Sachan, M., & Schölkopf, B. (2023). CLadder: Assessing causal reasoning in language models. Advances in Neural Information Processing Systems, 36. https://arxiv.org/abs/2312.04350

Kahneman, D. (2011). *Thinking, fast and slow*. Farrar, Straus and Giroux.

Kaplan, J., McCandlish, S., Henighan, T., Brown, T. B., Chess, B., Child, R., Gray, S., Radford, A., Wu, J., & Amodei, D. (2020). Scaling laws for neural language models. arXiv:2001.08361. https://arxiv.org/abs/2001.08361

Kim, M. J., Pertsch, K., Karamcheti, S., Xiao, T., Balakrishna, A., Nair, S., Rafailov, R., Foster, E., Lam, G., Sanketi, P., Vuong, Q., Kollar, T., Burchfiel, B., Tedrake, R.,

Sadigh, D., Levine, S., Liang, P., & Finn, C. (2024). OpenVLA: An open-source vision-language-action model. arXiv:2406.09246. https://arxiv.org/abs/2406.09246

Lake, B. M., Ullman, T. D., Tenenbaum, J. B., & Gershman, S. J. (2017). Building machines that learn and think like people. Behavioral and Brain Sciences, 40, e253. https://doi.org/10.1017/S0140525X16001837

Lakoff, G., & Johnson, M. (1999). *Philosophy in the flesh: The embodied mind and its challenge to Western thought*. Basic Books.

Le Guin, U. K. (1974). *The dispossessed: An ambiguous utopia*. Harper & Row.

Legg, S., & Hutter, M. (2007). Universal intelligence: A definition of machine intelligence. *Minds and Machines*, 17(4), 391-444. https://doi.org/10.1007/s11023-007-9079-x

Lucas, R. E. (1976). Econometric policy evaluation: A critique. *Carnegie-Rochester Conference Series on Public Policy*, 1, 19-46. https://doi.org/10.1016/S0167-2231(76)80003-6

Mancoridis, M., Weeks, B., Vafa, K., & Mullainathan, S. (2025). Potemkin understanding in large language models. arXiv:2506.21521. https://arxiv.org/abs/2506.21521

Marcus, G., & Davis, E. (2019). *Rebooting AI: Building artificial intelligence we can trust*. Pantheon Books.

Marr, D. (1982). Vision: A computational investigation into the human representation and processing of visual information. W.H. Freeman.

Metzinger, T. (2003). *Being no one: The self-model theory of subjectivity*. MIT Press.

Miéville, C. (2011). Embassytown. Macmillan.

Morris, M. R., Sohl-Dickstein, J., Fiedel, N., Warkentin, T., Dafoe, A., Faust, A., Farabet, C., & Legg, S. (2024). Position: Levels of AGI for operationalizing progress on the path to AGI. *Proceedings of the 41st International Conference on Machine Learning*, PMLR 235, 36308-36321. https://proceedings.mlr.press/v235/morris24b.html

Myerson, R. B. (1979). Incentive compatibility and the bargaining problem. *Econometrica*, 47(1), 61-73. https://doi.org/10.2307/1912346

Myerson, R. B. (1981). Optimal auction design. *Mathematics of Operations Research*, 6(1), 58-73. https://doi.org/10.1287/moor.6.1.58

Nagel, T. (1974). What is it like to be a bat? *Philosophical Review*, 83(4), 435-450. https://doi.org/10.2307/2183914

Newell, A., & Simon, H. A. (1976). Computer science as empirical inquiry: Symbols and search. Communications of the ACM, 19(3), 113-126. https://doi.org/10.1145/360018.360022

North, D. C. (1990). *Institutions, institutional change and economic performance*. Cambridge University Press.

Olson, E. T. (2023). Personal identity. In E. N. Zalta & U. Nodelman (Eds.), *Stanford encyclopedia of philosophy*. Stanford University. https://plato.stanford.edu/entries/identity-personal/

Packer, C., Wooders, S., Lin, K., Fang, V., Patil, S. G., Stoica, I., & Gonzalez, J. E. (2023). MemGPT: Towards LLMs as operating systems. arXiv:2310.08560. https://arxiv.org/abs/2310.08560

Parfit, D. (1984). *Reasons and persons*. Oxford University Press.

Pearl, J. (2000). *Causality: Models, reasoning, and inference*. Cambridge University Press.

Pearl, J. (2019). The seven tools of causal inference, with reflections on machine learning. Communications of the ACM, 62(3), 54-60. https://doi.org/10.1145/3241036

Pearl, J., & Mackenzie, D. (2018). *The book of why: The new science of cause and effect*. Basic Books.

Peng, X. B., Andrychowicz, M., Zaremba, W., & Abbeel, P. (2018). Sim-to-real transfer of robotic control with dynamics randomization. *2018 IEEE International Conference on Robotics and Automation (ICRA)*, pp. 3803-3810. https://doi.org/10.1109/ICRA.2018.8460528

Penn, D. C., Holyoak, K. J., & Povinelli, D. J. (2008). Darwin's mistake: Explaining the discontinuity between human and nonhuman minds. *Behavioral and Brain Sciences*, 31(2), 109-130. https://doi.org/10.1017/S0140525X08003543

Piaget, J. (1952). *The origins of intelligence in children*. International Universities Press.

Pohl, F., & Kornbluth, C. M. (1953). *The space merchants*. Ballantine Books.

Popper, K. R. (1959). *The logic of scientific discovery*. Hutchinson.

Premack, D., & Woodruff, G. (1978). Does the chimpanzee have a theory of mind? *Behavioral and Brain Sciences*, 1(4), 515-526. https://doi.org/10.1017/S0140525X00076512

Ray, S. (1965). Professor Shonku o Golok Rahasya [Professor Shonku and the Mystery of the Sphere]. *Sandesh*. Collected in *Professor Shonku*. NewScript Publications.

Reichenbach, H. (1938). *Experience and prediction: An analysis of the foundations and the structure of knowledge*. University of Chicago Press.

Rissanen, J. (1978). Modeling by shortest data descriptions. *Automatica*, 14(5), 465-471. https://doi.org/10.1016/0005-1098(78)90005-5

Schmidhuber, J. (1991). A possibility for implementing curiosity and boredom in model-building neural controllers. In J. A. Meyer & S. W. Wilson (Eds.), Proceedings of the First International Conference on Simulation of Adaptive Behavior (pp. 222-227). MIT Press. https://dl.acm.org/doi/10.5555/116517.116542

Schmidhuber, J. (2010). Formal theory of creativity, fun, and intrinsic motivation. *IEEE Transactions on Autonomous Mental Development*, 2(3), 230-247. https://doi.org/10.1109/TAMD.2010.2056368

Schölkopf, B., Locatello, F., Bauer, S., Ke, N. R., Kalchbrenner, N., Goyal, A., & Bengio, Y. (2021). Toward causal representation learning. Proceedings of the IEEE, 109(5), 612-634. https://doi.org/10.1109/JPROC.2021.3058954

Scott, W. R. (2014). *Institutions and organizations: Ideas, interests, and identities* (4th ed.). Sage.

Searle, J. R. (1980). Minds, brains, and programs. *Behavioral and Brain Sciences*, 3(3), 417-424. https://doi.org/10.1017/S0140525X00005756

Shapira, N., Levy, M., Alavi, S. H., Zhou, X., Choi, Y., Goldberg, Y., Sap, M., & Shwartz, V. (2024). Clever Hans or neural theory of mind? Stress testing social reasoning in large language models. In Proceedings of the 18th Conference of the European Chapter of the Association for Computational Linguistics (Volume 1:

Long Papers) (pp. 2257-2273). Association for Computational Linguistics. https://aclanthology.org/2024.eacl-long.138/

Shirow, M. (1989). *Ghost in the shell* [Manga series]. Kodansha.

Solomonoff, R. J. (1964). A formal theory of inductive inference. Part I. Information and Control, 7(1), 1-22. https://doi.org/10.1016/S0019-9958(64)90223-2

Sorensen, R. A. (1992). *Thought experiments*. Oxford University Press.

Srivastava, A., Rastogi, A., Rao, A., Shoeb, A. A. M., Abid, A., Fisch, A., Brown, A. R., Santoro, A., Gupta, A., Garriga-Alonso, A., Kluska, A., Lewkowycz, A., Agarwal, A., Power, A., Ray, A., Warstadt, A., Kocurek, A. W., Safaya, A., Tazarv, A., . . . Wu, Z. (2022). Beyond the imitation game: Quantifying and extrapolating the capabilities of language models. Transactions on Machine Learning Research. https://openreview.net/forum?id=uyTL5Bvosj

Strachan, J. W. A., Albergo, D., Borghini, G., Pansardi, O., Scaliti, E., Gupta, S., Saxena, K., Rufo, A., Panzeri, S., Manzi, G., Graziano, M. S. A., & Becchio, C. (2024). Testing theory of mind in large language models and humans. *Nature Human Behaviour*, 8(7), 1285-1295. https://doi.org/10.1038/s41562-024-01882-z

Street, W., Siy, J. O., Keeling, G., Baranes, A., Barnett, S., McKibben, M., Kanyere, T., Lentz, A., Aguera y Arcas, B., & Dunbar, R. I. M. (2025). LLMs achieve adult human performance on higher-order theory of mind tasks. Frontiers in Human Neuroscience, 19, 1633272. https://doi.org/10.3389/fnhum.2025.1633272

Sutton, R. (2019). The bitter lesson. *Incomplete Ideas* (blog). http://www.incompleteideas.net/IncIdeas/BitterLesson.html

Suvin, D. (1979). *Metamorphoses of science fiction: On the poetics and history of a literary genre*. Yale University Press.

Tchaikovsky, A. (2015). *Children of time*. Tor UK.

Tomasello, M. (1999). *The cultural origins of human cognition*. Harvard University Press.

Turing, A. M. (1950). Computing machinery and intelligence. Mind, 59(236), 433-460. https://doi.org/10.1093/mind/LIX.236.433

Vaswani, A., Shazeer, N., Parmar, N., Uszkoreit, J., Jones, L., Gomez, A. N., Kaiser, L., & Polosukhin, I. (2017). Attention is all you need. *Advances in Neural Information Processing Systems*, 30. https://arxiv.org/abs/1706.03762

Wei, J., Tay, Y., Bommasani, R., Raffel, C., Zoph, B., Borgeaud, S., Yogatama, D., Bosma, M., Zhou, D., Metzler, D., Chi, E. H., Hashimoto, T., Vinyals, O., Liang, P., Dean, J., & Fedus, W. (2022). Emergent abilities of large language models. *Transactions on Machine Learning Research*. https://openreview.net/forum?id=yzkSU5zdwD

Weir, A. (2014). *The Martian*. Crown Publishers.

Williamson, O. E. (1985). *The economic institutions of capitalism: Firms, markets, relational contracting*. Free Press.

Wimmer, H., & Perner, J. (1983). Beliefs about beliefs: Representation and constraining function of wrong beliefs in young children's understanding of deception. Cognition, 13(1), 103-128. https://doi.org/10.1016/0010-0277(83)90004-5